\documentclass[letter,11pt]{article}

\usepackage{fullpage}

\usepackage[cmex10]{amsmath}
\usepackage{epsfig,epsf,psfrag,amssymb,amsfonts,latexsym,graphicx,bm,cite,xcolor,url,array}
\usepackage{fixltx2e}
\usepackage{cases}
\usepackage{verbatim}
\usepackage[mathscr]{eucal}
\usepackage{algorithm, algorithmic}
\usepackage{amsthm}
\usepackage{caption,subcaption}

\usepackage[parfill]{parskip}

\usepackage{appendix}

\newtheorem{lem}{Lemma}

\DeclareMathOperator{\Supp}{Supp}

\DeclareMathOperator{\rank}{Rank}

\def\tha{{\mbox{\tiny th}}}

\DeclareMathOperator{\Diag}{Diag}
\DeclareMathOperator{\Diagn}{Diag_{\mathit{n}}}

 \def\0{{\bf 0}}

\def\viz{{viz.,\ \/}}

\def\nn{\nonumber}

\def\qed{\hfill\hbox{${\vcenter{\vbox{
    \hrule height 0.4pt\hbox{\vrule width 0.4pt height 6pt
    \kern5pt\vrule width 0.4pt}\hrule height 0.4pt}}}$}}

\definecolor{myred}{rgb}{0.3,0.0,0.7}
\definecolor{dkg}{rgb}{0.1,0.7,0.2}
\definecolor{dkb}{rgb}{0.0,0.2,0.8}

\def\bfi{{\mathbf i}}

\def\Ebb{{\mathbb E}}

\def\Rbb{{\mathbb R}}

\def\tilB{{\widetilde{B}}}
\def\tilCC{{\widetilde{C}}}

\newcommand{\bprfof}{\begin{proof_of}}
\newcommand{\eprfof}{\end{proof_of}}
\newcommand{\bprf}{\begin{myproof}}
\newcommand{\eprf}{\end{myproof}}
\newcommand{\bp}{\begin{psfrags}}
\newcommand{\ep}{\end{psfrags}}
\newcommand{\bl}{\begin{lemma}}
\newcommand{\el}{\end{lemma}}
\newcommand{\bt}{\begin{theorem}}
\newcommand{\et}{\end{theorem}}
\newcommand{\bc}{\begin{center}}
\newcommand{\ec}{\end{center}}
\newcommand{\bi}{\begin{itemize}}
\newcommand{\ei}{\end{itemize}}
\newcommand{\ben}{\begin{enumerate}}
\newcommand{\een}{\end{enumerate}}
\newcommand{\bd}{\begin{definition}}
\newcommand{\ed}{\end{definition}}
\def\beq{\begin{equation}}
\def\eeq{\end{equation}\noindent}
\def\beqn{\begin{eqnarray}}
\def\eeqn{\end{eqnarray} \noindent}
\def\beqnn{  \begin{eqnarray*}}
\def\eeqnn{\end{eqnarray*}  \noindent}
\def\bcase{  \begin{numcases}}
\def\ecase{\end{numcases}   \noindent}
\def\bsbcase{  \begin{subnumcases}}
\def\esbcase{\end{subnumcases}   \noindent}

\newtheorem{theorem}{Theorem}

\newtheorem{lemma}{Lemma}

\newtheorem{claim}{Claim}

\newtheorem{definition}{Definition}

\newtheorem{remark}{Remark}

\newtheorem{condition}{Condition}

\newenvironment{myproof}{\noindent{\em Proof:} \hspace*{1em}}{
    \hspace*{\fill} $\Box$ }
\newenvironment{proof_of}[1]{\noindent {\em Proof of #1: }}{\hspace*{\fill} $\Box$ }

\newcommand{\matplottc}[1]{               
        \unitlength .45truein
        \begin{center}
        \includegraphics{#1.ps}
        \end{picture}
        \end{center}
}

\def\psfancypar#1#2{\begingroup\def\par{\endgraf\endgroup\lineskiplimit=0pt}
               \setbox2=\hbox{\large\sc #2}
               \newdimen\tmpht \tmpht \ht2 \advance\tmpht by \baselineskip
               \font\hhuge=Times-Bold at \tmpht
               \setbox1=\hbox{{\hhuge #1}}
               \count7=\tmpht \count8=\ht1
               \divide\count8 by 1000 \divide\count7 by \count8
               \tmpht=.001\tmpht\multiply\tmpht by \count7
               \font\hhuge=Times-Bold at \tmpht
               \setbox1=\hbox{{\hhuge #1}}
               \noindent
                \hangindent1.05\wd1
               \hangafter=-2 {\hskip-\hangindent
               \lower1\ht1\hbox{\raise1.0\ht2\copy1}%
                \kern-0\wd1}\copy2\lineskiplimit=-1000pt}

\def\Kout{\setbox1=\hbox{\Huge\bf K}\hbox to
1.05\wd1{\hspace{.05\wd1}
}}
\def\Sout{\setbox1=\hbox{\Huge\bf S}\hbox to 1.05\wd1{\hspace{.05\wd1}
}}

\def\R{{\mathbb{R}}}

\allowdisplaybreaks[4]

\DeclareMathOperator{\vecform}{vec}

\DeclareMathOperator{\ten}{ten}
\DeclareMathOperator{\Col}{Col}

\DeclareMathOperator{\krank}{krank}
\def\ngram{{\odot n}}
\def\mgram{{\odot m}}

\def\Rest{{\operatorname{Rest.}}}

\title{When are Overcomplete Topic Models Identifiable?\\ Uniqueness of  Tensor Tucker Decompositions\\ with Structured Sparsity}

\author{Animashree Anandkumar, Daniel Hsu, Majid Janzamin and Sham Kakade\,\footnote{A. Anandkumar and M. Janzamin   are with the Center for Pervasive Communications and Computing, Electrical Engineering and Computer Science Dept., University of California, Irvine, USA 92697. Email: a.anandkumar@uci.edu,mjanzami@uci.edu. Daniel Hsu and Sham Kakade are with Microsoft Research New England, 1 Memorial Drive, Cambridge, MA 02142. Email: dahsu@microsoft.com, skakade@microsoft.com}}

\begin{document}

\maketitle

\begin{abstract}


Overcomplete latent representations have been very popular for unsupervised feature learning in recent years. In this paper, we specify which overcomplete models can be identified given observable moments of a certain order.
We consider   probabilistic admixture or topic models  in the overcomplete regime, where the number of latent topics can greatly exceed the size of the observed word vocabulary. While   general  overcomplete topic models are not   identifiable, we establish {\em generic} identifiability under  a  constraint, referred to as  {\em topic persistence}. Our sufficient conditions for identifiability involve a novel set of ``higher order'' expansion conditions on   the {\em topic-word matrix} or the {\em population structure}   of the   model. This set of higher-order expansion conditions allow for overcomplete models, and require  the existence of a perfect  matching    from latent topics to higher order observed words. We establish that random structured topic models
are identifiable w.h.p. in the overcomplete regime. Our identifiability results allows for   general (non-degenerate) distributions for modeling the topic proportions, and thus, we can handle arbitrarily correlated topics in our framework.  Our identifiability results imply uniqueness of a class of tensor decompositions with structured sparsity which is contained in the class of {\em Tucker} decompositions, but is more general than the {\em Candecomp/Parafac} (CP)   decomposition.
\end{abstract}






\section{Introduction}
The performance of   many machine learning methods is hugely dependent on the choice of data representations or features. Overcomplete representations,  where the number of features can be greater than the
dimensionality of the input data, have been extensively employed, and are arguably critical in a   number of applications such as speech and computer vision~\cite{bengio2012unsupervised}. Overcomplete representations are known to be more robust to noise, and can provide greater flexibility in modeling~\cite{Lewicki98learningovercomplete}. Unsupervised  estimation of overcomplete representations has been hugely popular  due to the availability of large-scale unlabeled samples in many applications.

A probabilistic framework for incorporating features  posits latent or hidden variables that can provide a good explanation to the observed data.
Overcomplete probabilistic models can incorporate a much larger number of latent variables compared to the  observed dimensionality.
In this paper, we characterize the conditions under which overcomplete latent variable models can be identified from their observed moments.

For any parametric statistical model, identifiability is a fundamental  question of whether the model parameters  can be uniquely recovered given the observed statistics. Identifiability is crucial in a number of applications where the latent variables are the quantities of interest, e.g. inferring  diseases (latent variables) through  symptoms (observations), inferring communities (latent variables) via the interactions among the actors in a social network (observations), and so on. Moreover, identifiability can be relevant even in  predictive settings, where feature learning is employed for some  higher level task such as classification. For instance, non-identifiability can lead to the presence of non-isolated local optima for optimization-based learning methods, and this can affect their convergence properties, e.g. see~\cite{uschmajew2012local}.

In this paper, we characterize identifiability for a popular class of   latent variable models, known as the  {\em admixture} or {\em topic} models~\cite{BleiNgJordan2003,PritchardEtal2000}. These are hierarchical mixture models, which incorporate the presence of multiple latent states (i.e. topics) in each document consisting of a tuple of observed variables (i.e. words). Previous works have established that the model parameters can be estimated efficiently using low order observed moments (second and third order) under some non-degeneracy assumptions, e.g.~\cite{AnandkumarEtal:tensor12,AnandkumarEtal:LinearBayesianLatent,AroraICML}. However, these non-degeneracy conditions imply that  the model is undercomplete, i.e., the latent dimensionality (number of topics)  cannot exceed the observed dimensionality (word vocabulary size).
In this paper, we remove this restriction and consider overcomplete topic models, where the number of topics can far exceed the word vocabulary size.

It is perhaps not surprising that general topic models are not identifiable in the overcomplete regime.
To this end, we introduce an additional constraint on the model, referred to as {\em topic persistence}. Intuitively, this captures the ``locality'' effect among the observed words, and is not present in the usual ``bag-of-words'' or {\em exchangeable} topic model. Such local dependencies among observations abound in applications  such as text, images and speech, and can lead to a more faithful representation. In addition, we establish that the presence of topic persistence is central towards obtaining model identifiability in the overcomplete regime, and we provide an in-depth analysis of this phenomenon in this paper.





\subsection{Summary of results}

In this paper, we  provide conditions for {\em generic}\,\footnote{A model is generically identifiable, if all the parameters in the parameter space are identifiable, almost surely. Refer to Definition \ref{def:generic ident.} for more discussion.} model identifiability   of  overcomplete topic models    given observable moments of a certain order (i.e., having a certain number of words in each document). We introduce the notion of {\em topic persistence}, and analyze its effect on identifiability. We establish  identifiability in the presence of a novel combinatorial object, referred to as {\em perfect $n$-gram matching}, in the bipartite graph from   topics to   words. Finally, we prove that random structured topic models satisfy these criteria, and are thus identifiable in the overcomplete regime.

\begin{figure} \centering
\bp\psfrag{h}[l]{\footnotesize$h$}
\psfrag{y1}[l]{\scriptsize$y_1$}
\psfrag{y2}[l]{\scriptsize$y_2$}
\psfrag{y2r}[l]{\scriptsize$y_{2r}$}
\psfrag{x1}[l]{\scriptsize$x_1$}
\psfrag{xn}[l]{\scriptsize$x_n$}
\psfrag{xn1}[l]{\scriptsize$x_{n\!+\!1}$}
\psfrag{xnn}[l]{\scriptsize$x_{2n}$}
\psfrag{x2rn1}[l]{\scriptsize$x_{(\!2r\!-\!1\!)\!n\!+\!1}$}
\psfrag{x2rn}[l]{\scriptsize$x_{2rn}$}
\psfrag{sa}[l]{$x_2$}
\psfrag{A}[l]{\scriptsize$A$}
\centering \includegraphics[width=2.5in]{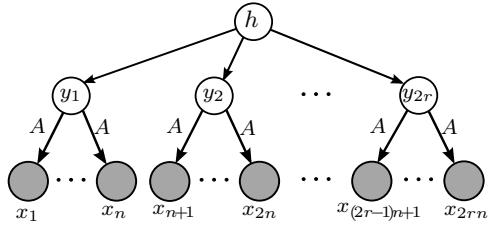}\ep
\caption{\small Hierarchical structure of the $n$-persistent topic model. $2rn$ number of words (views) are shown for some integer $r \geq 1$. A single topic $y_j, j \in [2r]$, is chosen for each $n$ successive views $\{ x_{(j-1)n+1},\dotsc,x_{(j-1)n+n} \}$. Matrix $A$ is the population structure or topic-word matrix.}
\label{fig:MultiViewModelProposed}
\end{figure}

\paragraph{Persistent Topic Model: }We first introduce the $n$-persistent topic model, where the parameter $n$ determines the  persistence level of a common topic in  a  sequence of $n$ successive words. For instance,   in Figure \ref{fig:MultiViewModelProposed}, the sequence of successive words $x_1, \ldots, x_n$ share a common topic $y_1$, and similarly, the words $x_{n+1},\ldots, x_{2n}$ share topic $y_2$, and so on. The $n$-persistent model reduces to the popular  ``bag-of-words'' model, when $n=1$, and to the single topic model (i.e. only one topic in each document) when $n \to \infty$. Intuitively,   topic persistence aids identifiability  since we have  multiple {\em views} of the common hidden  topic generating a sequence of successive words.
We establish that the   bag-of-words  model (with $n=1$) is too non-informative about the topics  in the overcomplete regime, and is therefore, not identifiable. On the other hand, $n$-persistent overcomplete topic models with $n\geq 2$ can become  identifiable, and we establish a set of transparent conditions for identifiability.

\paragraph{Deterministic Conditions for Identifiability: }Our sufficient conditions for identifiability are in the form of expansion conditions from the latent topic space to the observed word space. In the overcomplete regime, there are more topics than words in the vocabulary, and thus it is impossible to have expansion on the bipartite graph  from topics to  words, i.e., the graph encoding the sparsity pattern of the topic-word matrix. Instead, we impose an  expansion constraint   from topics to ``higher order'' words, which allows us to incorporate overcomplete models. We establish that this condition translates to the presence of a novel combinatorial object, referred to as the {\em perfect n-gram  matching}, on the topic-word bipartite graph. Intuitively, the perfect $n$-gram  matching condition   implies ``diversity'' among the   higher-order word supports for different topics which leads to identifiability.
In addition, we present trade-offs among   the following quantities: number of topics, size of the word vocabulary, the topic persistence level,   the order of the observed moments at hand,  the minimum and  maximum degrees of any topic in the topic-word bipartite graph, and the {\em Kruskal rank}~\cite{Kruskal:76} of the topic-word matrix,   under which identifiability holds. To the best of our knowledge, this is the first work to provide  conditions for characterizing identifiability of overcomplete topic models with structured sparsity.

\paragraph{Identifiability of Random Structured Topic Models: }We explicitly characterize the regime  of identifiability for the random setting, where each topic $i$ is randomly supported on a  set of  $d_i$ words, i.e. the bipartite graph is a random  graph.   For this  random model with $q$ topics,  $p$-dimensional word vocabulary, and topic persistence level $n$,     when $q = O(p^n)$ and
$\Theta(\log p )\leq d_i \leq \Theta( p^{1/n})$, for all topics $i$, the topic-word matrix  is identifiable from $2n^{\tha}$ order observed moments with high probability. Intuitively, the upper bound on the degrees $d_i$ is needed to limit the  overlap of word supports among different topics in the overcomplete regime: as the number of topics $q$ increases (i.e., $n$ increases in the above degree bound), the degree needs to be correspondingly smaller to ensure identifiability, and we make this dependence explicit. Intuitively, as the extent of overcompleteness increases, we need sparser connections from topics to words to ensure sufficient diversity in the word supports among different topics. The lower bound on the degrees is required so that there are enough edges in the topic-word bipartite graph so that various topics can be distinguished from one another. Furthermore, we establish that the size condition $q = O(p^n)$  for identifiability is tight.



\paragraph{Implications on Uniqueness of Overcomplete  Tucker and CP Tensor Decompositions: }We establish that identifiability of an overcomplete topic model is equivalent  to uniqueness of decomposition of the observed moment tensor (of a certain order).
Our identifiability results for persistent topic models   imply uniqueness of a structured class of tensor decompositions, which is contained in the class of {\em Tucker} decompositions, but is more general than  the {\em candecomp/parafac} (CP) decomposition~\cite{kolda_survey}. This sub-class of Tucker decompositions involves structured sparsity and symmetry constraints on the {\em core tensor}, and sparsity constraints on the {\em inverse factors} of the Tucker decomposition. The structural constraints on the Tucker tensor decomposition are related to  the topic model as follows: the sparsity and symmetry constraints on the core tensor are related   to the persistence property of the topic model, and the sparsity constraints on  the inverse factors are equivalent to the sparsity constraints on the topic-word matrix.
For   $n$-persistent topic model with  $n=1$ (bag-of-words   model),  the tensor decomposition is a general Tucker decomposition, where the core tensor is fully dense, while for   $n\to \infty$ (single-topic model), the tensor decomposition reduces to a CP decomposition, i.e. the core tensor is a {\em diagonal tensor}. For a finite persistence level $n$, in between these two extremes, the core tensor satisfies certain  sparsity and symmetry constraints, which becomes  crucial towards  establishing identifiability in the overcomplete regime.

\subsection{Overview of Techniques}

We now provide a short overview of the techniques employed in this paper.

\paragraph{Recap of Identifiability Conditions in   Under-complete Setting (Expansion Conditions on Topic-Word Matrix): }Our approach is based on the recent results of~\cite{AnandkumarEtal:LinearBayesianLatent}, where conditions for  identifiability  of topic models are derived,   given  pairwise observed moments (specifically, co-occurrence of word-pairs in documents). Consider a topic model with $q$ topics  and observed word vocabulary of size $p$. Let $A\in \Rbb^{p \times q}$ denote the topic-word matrix.
Expansion conditions are imposed in~\cite{AnandkumarEtal:LinearBayesianLatent} on the topic-word bipartite graph which  imply that  (generically) the sparsest vectors in the column span of  $A$, denoted by Col$(A)$, are the columns of $A$ themselves.  Thus the topic-word matrix $A$ is identifiable from pairwise moments under expansion constraints. However,  these expansion conditions constrain the model to be    under-complete, i.e., the number of topics $q\leq p$, the size of the word vocabulary. Therefore, the techniques derived in~\cite{AnandkumarEtal:LinearBayesianLatent} are not directly applicable here since we consider overcomplete models.

\paragraph{Identifiability in Overcomplete Setting and Why Topic-Persistence Helps: } Pairwise moments are thus not sufficient for identifiability of overcomplete models, and the question is whether higher order moments can yield identifiability. We can  view the  higher order moments  as  pairwise moments of another equivalent topic model, which enables us to apply the techniques of~\cite{AnandkumarEtal:LinearBayesianLatent}.
The key question is whether we have expansion in the equivalent topic model, which implies identifiability. For a general topic model (without any topic persistence constraints), it can be shown that for identifiability, we require expansion of the $n^{\tha}$-order {\em Kronecker product} of the original topic-word matrix $A$, denoted by $A^{\otimes n}\in \Rbb^{p^n\times q^n}$, when given access to $(2n)^{\tha}$-order moments, for any integer $n \geq 1$. In the overcomplete regime where $q > p$,   $A^{\otimes n}$ cannot expand, and therefore, overcomplete models are not identifiable in general. On the other hand, we show that imposing the constraint of topic persistence can lead to identifiability.  For a $n$-persistent topic model, given $(2n)^{\tha}$-order moments, we establish that identifiability occurs when the $n^{\tha}$-order {\em Khatri-Rao product} of $A$, denoted by $A^{\odot n}\in \Rbb^{p^n\times q}$, expands. Note that the Khatri-Rao product $A^{\odot n}$ is a sub-matrix of the Kronecker product $A^{\otimes n}$, and the Khatri-Rao product $A^{\odot n}$ can expand as long as $q \leq p^n$.
Thus, the property of topic persistence is central towards achieving identifiability in the overcomplete regime.

\paragraph{First-Order Approach for Identifiability of   Overcomplete Models (Expansion of $n$-gram Topic-Word Matrix): }We refer to $A^\ngram\in \Rbb^{p^n \times q}$ as the $n$-gram topic-word matrix, and intuitively, it relates topics to $n$-tuple words. Imposing  the expansion conditions derived in~\cite{AnandkumarEtal:LinearBayesianLatent} on  $A^\ngram$ implies   that (generically) the sparsest vectors in   Col$(A^\ngram)$, are the columns of $A^\ngram$ themselves. Thus,  the topic-word matrix $A$
is identifiable from $(2n)^{\tha}$-order moments for a $n$-persistent topic model.  We refer to this as the ``first-order'' approach since we directly impose the expansion conditions of~\cite{AnandkumarEtal:LinearBayesianLatent} on $A^\ngram$, without exploiting the additional structure present in $A^\ngram$.

\paragraph{Why the First-Order Approach is not Enough: }Note that $A^\ngram\in \Rbb^{p^n \times q}$ matrix relates topics to $n$-tuples of words. Thus, the entries of  $A^\ngram$   are highly correlated, even if the original topic-word matrix $A$ is assumed to be randomly generated. It is non-trivial to derive conditions on $A$, so that $A^\ngram$  expands.
Moreover, we establish that   $A^\ngram$ fails to expand on ``small'' sets, as required in~\cite{AnandkumarEtal:LinearBayesianLatent}, when the degrees are sufficiently different\,\footnote{For $A^{\ngram}$ to expand  on a set of size $s\geq 2$, it is necessary that  $s\cdot {d_{\min}+n-1 \choose n} \geq s+ {d_{\max}+n-1 \choose n}$, where $d_{\min}$ and $d_{\max}$ are the minimum and maximum degrees, and $n$ is the extent of overcompleteness: $q= \Theta(p^n)$. When   the model is highly overcomplete (large $n$) and  we require small set expansion (small $s$), the degrees need to be nearly the same. Thus, it is desirable to impose expansion only on large sets, since it allows for more degree diversity.}. Thus, the first-order approach is highly restrictive   in the overcomplete setting.

\paragraph{Incorporating Rank Criterion: }Note that   $A^\ngram$   is highly structured: the columns of $A^\ngram$ matrix possess a tensor\,\footnote{When any column of  $A^\ngram\in \Rbb^{p^n \times q}$ (of length $p^n$) is reshaped as a $n^{\tha}$-order tensor $T\in \Rbb^{p \times p \times \dotsb \times p}$, the tensor $T$ is rank $1$.} rank of $1$, when $n > 1$. This can be incorporated in our identifiability criteria as follows: we provide conditions under which the sparsest  vectors in Col$(A^\ngram)$, which also possess a tensor rank of $1$, are the columns of $A^\ngram$ themselves. This implies  identifiability of a $n$-persistent topic model, when given access to $(2n)^{\tha}$-order moments.
Note that
when a small number of columns of $A^\ngram$ are combined, the resulting vector cannot possess a tensor rank of $1$, and thus, we can rule out that such sparse combinations of columns using the rank criterion. The maximum such number is at least the {\em Kruskal rank}\,\footnote{The Kruskal rank is the maximum number  $k$ such that every $k$-subset of columns of $A$ are linearly independent. Note that the Kruskal rank is equal to the rank of $A$, when $A$ has full column rank. But this cannot happen in the overcomplete setting.} of $A$. Thus,  sparse combinations of columns of $A$ (up to the Kruskal rank) can be ruled out using  the rank criterion, and we require expansion on $A^\ngram$   only on large sets of topics (of size larger than the Kruskal rank). This agrees with the intuition that when the topic-word matrix $A$ has a larger Kruskal rank, it should be easier to identify $A$, since the Kruskal rank is related to the {\em mutual incoherence}\,\footnote{It is easy to show that   $\krank\geq (\max_{i \neq j} |a_i^\top a_j|)^{-1}$, where $a_i,a_j $ are  any pair of columns of $A$. Thus, higher incoherence leads to a larger kruskal rank.} among the columns of $A$, see~\cite{gandy2011tensor}.

\paragraph{Notion of Perfect $n$-gram Matching and Final Identifiability Conditions: }
Thus, we establish identifiability of overcomplete topic models subject to expansion  conditions $A^\ngram$ on sets of size larger than the  Kruskal rank of the topic-word matrix $A$. However, it is desirable to impose transparent and interpretable conditions directly on $A$ for identifiability. We introduce the notion of  {\em perfect $n$-gram matching}
on the topic-word bipartite graph, which ensures that each topic can be uniquely matched to a $n$-tuple word. This combined with a lower bound on the Kruskal rank provides the final set of deterministic conditions for identifiability of the overcomplete topic model. Intuitively, we require that  the columns of $A$ be sparse, while still maintaining a large enough Kruskal rank; in other words, the topics have to be sparse and  have sufficiently diverse word supports. Thus, we establish identifiability under a set of transparent conditions on the topic-word matrix $A$, consisting of perfect $n$-gram matching condition and a lower bound on the Kruskal rank of $A$.

\paragraph{Analysis under Random-Structured Topic-Word Matrices: }
Finally, we establish that the derived deterministic conditions are satisfied when the topic-word bipartite graph is randomly generated, as long as the degrees satisfy certain lower and upper bounds. Intuitively, a lower bound on the degrees of the topics is required to have degree concentration on various subsets so that expansion can occur, while the upper bound is required so that the Kruskal rank of the topic-word matrix is large enough compared to the sparsity level. Here, the main technical result is establishing the presence of a perfect $n$-gram matching in a random bipartite graph with a wide range of degrees. We present a greedy and a recursive  mechanism for constructing such a $n$-gram matching  for overcomplete models, which can be relevant even in other settings. For instance, our results imply the presence of a perfect matching when the edges of a bipartite graph are correlated in a structured manner, as given by the Khatri-Rao product.





\subsection{Related works}\label{sec:related}

We now summarize some recent related works in the area of identifiability and learning of latent variable models.

\paragraph{Identifiability, learning and applications of overcomplete latent representations:}
Many recent works employ unsupervised estimation of overcomplete features for higher level tasks such classification, e.g.~\cite{CoatesNgLee11,LeEtal:2011NIPS,dengdeep,bengio2012unsupervised}, and record huge gains over other approaches in a number of applications such as speech recognition and computer vision. However,     theoretical understanding regarding learnability or identifiability  of overcomplete representations is far more limited.

Overcomplete latent representations have been analyzed in the context of the independent components analysis (ICA), where the sources are assumed to be independent, and the mixing matrix is unknown. In the overcomplete or under-determined regime of the ICA, there are more sources than sensors. Identifiability and learning of the overcomplete ICA reduces to the problem of finding an overcomplete candecomp/parafac (CP) tensor decomposition. The classical result by Kruskal provides conditions for uniqueness of a CP decomposition~\cite{Kruskal:76,Kruskal:77}, with recent extensions to the notion of robust identifiability~\cite{Kruskalrobust}.
These results provide conditions for strict identifiability of the model, and here, the dimensionality of the latent space is required   to be of the same order as the observed space dimensionality. In contrast,  a number of recent works   analyze {\em generic} identifiability of overcomplete  CP decomposition, which is weaker than strict identifiability, e.g.~
\cite{jiang2004kruskal,LathauwerSIAM2006,STEGEMAN2006:Psychometrika,DeLathauwerEtal:FOOBI,chiantini2012generic,bocci2013refined,chiantini2013one}.
These works assume that the factors (i.e. the components) of the CP decomposition are generically drawn and provide conditions for uniqueness. They allow for  the latent dimensionality to be much larger (polynomially larger) than the observed dimensionality. These results
on the uniqueness of CP   decompositions also lead to identifiability of other latent variable models, such as  latent tree models, e.g.~\cite{Allman:09Stat,Allmanetal:12Arxiv}, and the single-topic model, or more generally latent Dirichlet allocation (LDA). Recently Goyal~et. al.~\cite{FourierPCA} proposed an alternative framework for overcomplete ICA models based on the eigen-decomposition of the reweighted covariance matrix (or higher order moments), where the weights are the Fourier coefficients. However, their approach  requires independence of sources (i.e. latent topics in our context), which is not imposed here.

In contrast to the above works dealing with the  CP tensor decomposition, we require  uniqueness for a more general class of tensor decompositions, in order to establish identifiability of topic models  with arbitrarily correlated topics.
We   establish that our class of tensor decomposition is contained in the class of {\em Tucker} decompositions which is more general than CP decomposition.
Moreover, we explicitly characterize the effect of the sparsity pattern of the factors (i.e., the topic-word matrix) on model identifiability, while all the previous works based on generic identifiability assume fully dense   factors (since sparse factors are not generic).
For a general overview of  tensor decompositions, see~\cite{kolda_survey,landsberg2012tensors}.




\paragraph{Identifiability and learning of undercomplete/over-determined latent representations: }Much of the theoretical results on identifiability and learning of the latent variable models are limited to non-singular models, which implies that the latent space dimensionality is at most the observed dimensionality. We outline some of the recent works below.

The works of Anandkumar et. al.~\cite{AnandkumarHsuKakade:COLT12,AnandkumarEtal:ldaNIPS12,AnandkumarEtal:tensor12} provide an efficient moment-based approach for learning topic models, under  constraints on the distribution of the topic proportions, e.g. the single topic model, and more generally  latent Dirichlet allocation (LDA).   In addition, the approach can handle a variety of latent variable models such as Gaussian mixtures, hidden Markov models (HMM) and community models~\cite{AnandkumarEtal:community12COLT}. The high-level idea is to reduce the problem of learning of the latent variable model to finding a CP decomposition of the (suitably adjusted) observed moment tensor. Various approaches can then be employed to find the CP decomposition. In~\cite{AnandkumarEtal:tensor12}, a tensor power method approach is analyzed and is shown to be an efficient guaranteed recovery method in the non-degenerate (i.e. undercomplete) setting. Previously, simultaneous diagonalization techniques have been employed for solving the CP decomposition, e.g.~\cite{AnandkumarHsuKakade:COLT12,mossel2005learning,chang1996full}. However, these techniques fail when the model is overcomplete, as considered here. We note that some recent techniques, e.g.~\cite{DeLathauwerEtal:FOOBI}, can be employed instead, albeit at a cost of higher computational complexity  for overcomplete CP tensor decomposition. However, it is not clear how the sparsity constraints affect the guarantees of such methods. Moreover, these approaches cannot handle general topic models, where the distribution of the topic proportions is not limited to these classes (i.e. either single topic or Dirichlet distribution), and we require tensor decompositions which are more general than the CP decomposition.

There are many other works which consider learning mixture models when multiple views are available. See~\cite{AnandkumarHsuKakade:COLT12} for a detailed description of these works. Recently,
Rabani et. al.~\cite{rabani2012learning} consider learning discrete mixtures given a large number of ``views'', and they refer to the number of views as the {\em sampling aperture}. They establish improved recovery results (in terms of $\ell_1$ bounds) when sufficient number of views are available ($2k-1$ views for a $k$-component mixture). However, their results are limited to discrete mixtures or single-topic models, while our setting can handle more general topic models.  Moreover, our approach is different since we incorporate sparsity constraints in the topic-word distribution. Another series of recent works by Arora et. al.~\cite{arora2012learning,AroraICML} employ approaches based on  non-negative matrix factorization (NMF) to recover the topic-word matrix. These works allow models with arbitrarily correlated topics, as considered here. They establish guaranteed learning when every topic has an {\em anchor} word, i.e. the word is uniquely generated from that topic, and does not occur under any other topic. Note that the anchor-word assumption cannot be satisfied in the overcomplete setting.

Our work is closely related to the work of Anandkumar et. al.~\cite{AnandkumarEtal:LinearBayesianLatent} which considers identifiability and learning of topic models   under expansion conditions on the topic-word matrix. The work of Spielman et. al~\cite{spielman2012exact} considers the problem of   dictionary learning, which is closely related to the setting of~\cite{AnandkumarEtal:LinearBayesianLatent}, but in addition assumes that the coefficient matrix is random. However, these works~\cite{AnandkumarEtal:LinearBayesianLatent,spielman2012exact} can handle only the under-complete setting, where the number of topics is less than the dimensionality of the word vocabulary (or the number of dictionary atoms is less than the number of observations in~\cite{spielman2012exact}). We extend these results to the overcomplete setting by proposing novel higher order expansion conditions on the topic-word matrix, and also incorporate additional rank constraints present in higher order moments.






\paragraph{Dictionary learning/sparse coding: }Overcomplete representations have been very popular in the context of dictionary learning or sparse coding. Here, the task is to jointly learn a dictionary as well as a sparse selection of the dictionary atoms to fit the observed data. There have been Bayesian as well as frequentist approaches for dictionary learning~\cite{Lewicki98learningovercomplete,KreutzEtal:DicLearning,Rao&Kreutz:FOCUSS}.
However, the heuristics employed in these works \cite{Lewicki98learningovercomplete,KreutzEtal:DicLearning,Rao&Kreutz:FOCUSS} have no performance guarantees. The work of Spielman et. al~\cite{spielman2012exact} considers learning (undercomplete) dictionaries and provide guaranteed learning under the assumption that the coefficient matrix is random (distributed as Bernoulli-Gaussian variables).
Recent works~\cite{MehtaGray:13ICML,MaurerEtal2012} provide generalization bounds for predictive sparse coding, where the goal of the learned representation is to obtain good performance on some predictive task. This differs from our framework since we do not consider predictive tasks here, but the task of recovering the underlying latent representation.  Hillar and Sommer~\cite{hillar2011ramsey} consider the problem of identifiability of sparse coding and establish that when the dictionary succeeds in reconstructing a certain set of sparse vectors, then there exists a unique sparse coding, up to permutation and scaling. However, our setting here is different, since we do not assume that a sparse set of topics occur in each document.

\section{Model} \label{sec:model}

\paragraph{Notation: }
The set $\{ 1,2,\dotsc,n \} $ is denoted by $[n] := \{ 1,2,\dotsc,n \} $. Given a set $X = \{1,\dotsc,p\}$, set $X^{(n)}$ denotes all ordered $n$-tuples generated from $X$. The cardinality of a set $S$ is denoted by $|S|$. For any vector $u$ (or matrix $U$), the support is denoted by $\Supp(u)$, and the $\ell_0$ norm is denoted by $\| u \|_0$, which corresponds to the number of non-zero entries of $u$, i.e., $\| u \|_0 := | \Supp(u)|$. For a vector $u \in \Rbb^q$, $\Diag(u) \in \Rbb^{q \times q}$ is the diagonal matrix with vector $u$ on its diagonal. The column space of a matrix $A$ is denoted by $\Col(A)$. Vector $e_i \in \mathbb{R}^q$ is the $i$-th basis vector, with the $i$-th entry equal to 1 and all the others equal to zero. 
For $A \in \Rbb^{p \times q}$ and $B \in \Rbb^{m \times n}$, the {\em Kronecker} product $A \otimes B \in \Rbb^{pm \times qn}$ is defined as \cite{Golub&VanLoan:book}
\begin{align*}
A \otimes B = \left[
\begin{array}{cccc}
a_{11} B & a_{12}B & \dotsb & a_{1q} B \\
a_{21} B & a_{22}B & \dotsb & a_{2q} B \\
\vdots & \vdots & \ddots & \vdots \\
a_{p1} B & a_{p2} B & \dotsb & a_{pq} B
\end{array} \right],
\end{align*}
and for $A = [a_1 | a_2 | \dotsb | a_r] \in \Rbb^{p \times r}$ and $B = [b_1 | b_2 | \dotsb | b_r] \in \Rbb^{m \times r}$, the {\em Khatri-Rao} product $A \odot B \in \Rbb^{pm \times r}$ is defined as
\begin{align*}
A \odot B = \left[ a_1 \otimes b_1 | a_2 \otimes b_2 | \dotsb | a_r \otimes b_r \right].
\end{align*}

\subsection{Persistent topic model} \label{sec:persistent topic model}

In this section, the {\em $n$-persistent topic model} is introduced and this imposes an additional constraint, known as topic persistence on the popular admixture model\cite{BleiNgJordan2003,PritchardEtal2000,nguyen2012posterior}. 
The $n$-persistent topic model reduces to the bag-of-words admixture model when $n=1$. 

An admixture model specifies a $q$-dimensional vector of topic proportions $h\in \Delta^{q-1} := \{u \in \mathbb{R}^q : u_i \geq 0, \sum_{i=1}^q u_i = 1\}$ which generates the observed variables $x_l \in \Rbb^p$ through vectors $a_1,\dotsc,a_q \in \Rbb^p$. This collection of vectors $a_i, i \in [q]$, is referred to as the {\em population structure} or the {\em topic-word matrix} \cite{nguyen2012posterior}. For instance, $a_i$  is the conditional distribution of words given topic $i$. The latent variable $h$ is a $q$ dimensional random vector $h := [h_1,\dotsc,h_q]^\top$ known as proportion vector. A prior distribution $P(h)$ over the probability simplex $\Delta^{q-1}$ characterizes the prior joint distribution over the latent variables $h_i, \ i \in [q]$. In the topic modeling, this is the prior distribution over the $q$ topics.

The $n$-persistent topic model has a three-level multi-view hierarchy in Figure \ref{fig:MultiViewModelProposed}. $2rn$ number of words (views) are shown in the model for some integer $r \geq 1$. In this model, a common hidden topic is persistent for a sequence of $n$ words $\{ x_{(j-1)n+1},\dotsc,x_{(j-1)n+n} \}, j \in [2r]$. 
Note that the random observed variables (words) are exchangeable within groups of size $n$, where $n$ is the persistence level, but are not globally exchangeable. 

We now describe  a linear representation of the $n$-persistent topic model, on lines of \cite{AnandkumarEtal:tensor12}, but with extensions to incorporate persistence.
Each random variable $y_j, j \in [2r],$ is a discrete valued random variable taking one of the $q$ possibilities $\{ 1,\dotsc,q \}$, i.e., $y_j \in [q]$ for $j \in [2r]$. In the $n$-persistent model, a single common topic is chosen for a sequence of $n$ words $\{ x_{(j-1)n+1},\dotsc,x_{(j-1)n+n} \}, j \in [2r]$, i.e., the topic is persistent for $n$ successive views.
For notational purposes, we equivalently assume that variables $y_j, j \in [2r]$, are encoded by the basis vectors $e_i, \ i \in [q]$. Thus, the variable $y_j, j \in [2r]$, is
\begin{align*}
y_j=e_i \in \Rbb^{q} \Longleftrightarrow \text{the topic of $j$-th group of words is $i$}.
\end{align*}
Given proportion vector $h$, topics $y_j, j \in [2r]$, are independently drawn according to the conditional expectation
\begin{align*}
\Ebb \bigl[ y_j | h \bigr] = h, \quad j \in [2r],
\end{align*}
or equivalently $\Pr \bigl[ y_j = e_i |h \bigr] = h_i, j \in [2r], i \in [q]$.

Finally, at the bottom layer, each observed variable $x_l$ for  $l \in [2rn]$, is a discrete-valued $p$-dimensional random variable, where $p$ is the size of word vocabulary. Again, we assume that variables $x_l$, are encoded by the basis vectors $e_k, \ k \in [p]$, such as 
\begin{align*}
x_l = e_k \in \Rbb^{p} \Longleftrightarrow \text{the $l$-th word in the document is $k$}.
\end{align*}
Given the corresponding topic $y_j, j \in [2r]$, words $x_l, l \in [2rn]$, are independently drawn according to the conditional expectation
\begin{align} \label{eq:moment propery of the persistent model}
\mathbb{E} \bigl[ x_{(j-1)n+k} | y_j = e_i \bigr] = a_i, \, i \in [q],  j \in [2r], \ k \in [n],
\end{align}
where vectors $a_i \in \mathbb{R}^p, \ i \in [q]$, are the conditional probability distribution vectors. The matrix $A=[a_1|a_2|\dotsb|a_q] \in \Rbb^{p \times q}$ collecting these vectors is the {\em population structure} or {\em topic-word matrix}.


The $(2rn)$-th order moment of observed variables $x_l, l \in [2rn]$, for some integer $r \geq 1$, is defined as (in the matrix form)\,\footnote{Vector $x$ is the vector generated by concatenating all vectors $x_l, l \in [2rn]$.}
\begin{align} \label{eq:4th order moment of x definition}
M_{2rn}(x) := \Ebb \left[ (x_1 \otimes x_2 \otimes \dotsm \otimes x_{rn})(x_{rn+1} \otimes x_{rn+2} \otimes \dotsm \otimes x_{2rn})^\top \right] \in \Rbb^{p^{rn} \times p^{rn}}.
\end{align}
For the $n$-persistent topic model with $2rn$ number of observations (words) $x_l, l \in [2rn]$, the corresponding moment is denoted by $M^{(n)}_{2rn}(x)$. Note that to estimate the $(2rn)^{\tha}$ moment, we require a minimum of $2rn$ words in each document. We can select the first $2rn$ words in each document, and average over the different documents to obtain a consistent estimate of the moment. In this paper, we consider the problem of identifiability when exact moments are available.\\
The moment characterization of the $n$-persistent topic model is provided in Lemma \ref{lem:persistent topic model moment characterization} in Section \ref{sec:persistent topic model moment characterization}. Given $M^{(n)}_{2rn}(x)$, what are the sufficient conditions under which the population structure $A$ is identifiable? This is answered in Section \ref{sec:ident. conditions}. 


\begin{remark}
Note that our results are valid for the more general linear model $x_l = A y_j$ (more precisely, $x_{(j-1)n+k} = A y_j, j \in [2r], k \in [n]$), i.e., each column of matrix $A$ does not need to be a valid probability distribution. Furthermore, the observed random variables $x_l$, can be continuous while the hidden ones $y_j$ are assumed to be discrete.
\end{remark}


\section{Sufficient  Conditions for Generic Identifiability} \label{sec:ident. conditions}
In this section, the identifiability result for the $n$-persistent topic model with access to $(2n)$-th order observed moment is provided. 
First, sufficient deterministic conditions on the population structure $A$ are provided for identifiability in Theorem \ref{Thm:Identifiability based on A}. Next, the deterministic analysis is specialized to a random structured model in Theorem \ref{Thm:Identifiability random}. 

We now make the notion of identifiability precise. As defined in literature, (strict) identifiability means that the population structure $A$ can be uniquely recovered up to permutation and scaling for all $A \in \Rbb^{p \times q}$.
Instead, we consider a more relaxed notion of identifiability, known as generic identifiability.
\begin{definition}[Generic identifiability] \label{def:generic ident.}
We refer to a matrix $A \in \Rbb^{p \times q}$ as generic, with a fixed sparsity pattern when the nonzero entries of $A$ are drawn from a distribution which is absolutely continuous with respect to Lebesgue measure\,\footnote{As an equivalent definition, if the non-zero entries of an arbitrary sparse matrix are independently perturbed with noise drawn from a continuous distribution to generate $A$, then $A$ is called generic.}.
For a given sparsity pattern, the class of population structure matrices is said to be {\em generically identifiable}~\cite{Allmanetal:12Arxiv}, if all the non-identifiable matrices form a set of Lebesgue measure zero.
\end{definition}

The $(2r)$-th order moment of hidden variables $h \in \Rbb^q$, denoted by $M_{2r}(h) \in \Rbb^{q^r \times q^r}$, is defined as
\begin{align} \label{eq:4th order moment of h}
M_{2r}(h) := \Ebb \biggl[  \Bigl( \overbrace{h \otimes \dotsm \otimes h}^{r \operatorname{times}} \Bigr)  \Bigr( \overbrace{h \otimes \dotsm \otimes h}^{r \operatorname{times}} \Bigl) ^\top \biggr] \in \Rbb^{q^r \times q^r}.
\end{align}

We now provide a set of sufficient conditions for generic identifiability of structured topic models given $(2rn)$-th order observed moment. We first start with a natural assumption on the hidden variables.


\begin{condition}[Non-degeneracy] \label{cond:non-degeneracy}
The $(2r)$-th order moment of hidden variables $h \in \Rbb^q$, defined in equation \eqref{eq:4th order moment of h}, is full rank (non-degeneracy of hidden nodes).
\end{condition}
Note that there is no hope of distinguishing distinct hidden nodes without this non-degeneracy assumption. We do not impose any other assumption on hidden variables and can incorporate arbitrarily correlated topics.

Furthermore, we can only hope to identify the population structure $A$ up to scaling and permutation. 
Therefore, we can identify $A$ up to a canonical form defined as:

\begin{definition}[Canonical form]
Population structure $A$ is said to be in \textnormal{canonical form} if all of its columns have unit norm.
\end{definition}

\subsection{Deterministic conditions for generic identifiability} \label{sec:ident. conditions deterministic}
In this section, we consider a fixed sparsity pattern on the population structure $A$ and establish generic identifiability when non-zero entries of $A$ are drawn from some continuous distribution.
Before providing the main result, a generalized notion of (perfect) matching for bipartite graphs is defined. 
We subsequently impose these conditions on the bipartite graph from topics to words which encodes the sparsity pattern of population structure $A$.

\subsubsection*{Generalized matching for bipartite graphs}
A bipartite graph with two disjoint vertex sets $Y$ and $X$ and an edge set $E$ between them is denoted by $G(Y,X;E)$.
Given the bi-adjacency matrix $A$, the notation $G(Y,X;A)$ is also used to denote a bipartite graph. Here, the rows and columns of matrix $A \in \Rbb^{|X| \times |Y|}$ are respectively indexed by $X$ and $Y$ vertex sets. 
For any subset $S \subseteq Y$, the set of neighbors of vertices in $S$ with respect to $A$ is defined as $N_A(S):= \{ i \in X : A_{ij} \neq 0 \ \operatorname{for} \ \operatorname{some} \ j \in S \}$, or equivalently, $N_E(S):= \{ i \in X : (j,i) \in E \ \operatorname{for} \ \operatorname{some} \ j \in S \}$ with respect to edge set $E$.


Here, we define a generalized notion of matching for a bipartite graph and refer to it as $n$-gram matching.
\begin{definition}[(Perfect) $n$-gram matching] \label{def:$n$-gram Matching}
A \textnormal{$n$-gram matching} $M$ for a bipartite graph $G(Y,X;E)$ is a subset of edges $M \subseteq E$ which satisfies the following conditions.
 First, for any $j \in Y$, we have $|N_M(j)| \leq n$. Second, for any $j_1,j_2 \in Y, j_1 \neq j_2$, we have $\min \{ |N_M(j_1)|,|N_M(j_2)| \} > | N_M(j_1) \cap N_M(j_2) |$. \\
A \textnormal{perfect $n$-gram matching} or \textnormal{$Y$-saturating $n$-gram matching} for the bipartite graph $G(Y,X;E)$ is a $n$-gram matching $M$ in which each vertex in $Y$ is the end-point of exactly $n$ edges in $M$.
\end{definition}

In words, in a $n$-gram matching $M$, each vertex $j \in Y$ is at most the end-point of $n$ edges in $M$ and for any pair of vertices in $Y$ ($j_1,j_2 \in Y, j_1 \neq j_2$), there exists at least one non-common neighbor in set $X$ for each of them ($j_1$ and $j_2$).

\begin{figure}\centering
\bp \psfrag{X}[l]{\small$Y$} \psfrag{Y}[l]{\small$X$}
\includegraphics[width=2.5in]{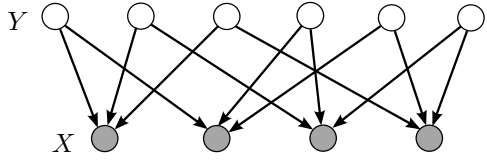} \ep
\caption{\small A bipartite graph $G(Y,X;E)$ with $|X|=4$ and $|Y|=6$ where the edge set $E$ itself is a perfect 2-gram matching.} \label{fig:perfect ngram example}
\end{figure}

As an example, a bipartite graph $G(Y,X;E)$ with $|X|=4$ and $|Y|=6$ is shown in Figure \ref{fig:perfect ngram example} for which the edge set $E$ itself is a perfect 2-gram matching.

\begin{remark}[Relationship to other matchings] The relationship of $n$-gram matching to other types of matchings is discussed below.
\bi
\item Regular matching: For special case $n=1$, the (perfect) $n$-gram matching reduces to the usual (perfect) matching for bipartite graphs.
\item $b$-matching: A $b$-matching for a bipartite graph $G(Y,X;E)$ (with equal vertex sizes $|X|=|Y|$) is a subset of edges $M_b \subseteq E$, where each vertex is connected to $b$ edges. Comparing with the proposed perfect $n$-gram matching, $b$-matching does not enforce that the set of neighbors be different, and furthermore, it requires that $X=Y$, which is not possible under the overcomplete setting.
\ei
\end{remark}


\begin{remark}[Necessary size bound] \label{remark:ngram perfect matching necessary bound}
Consider a bipartite graph $G(Y,X;E)$ with $|Y|=q$ and $|X|=p$ which has a perfect $n$-gram matching. 
Note that there are $p \choose n$ $n$-combinations on $X$ side and each combination can at most have one neighbor (a node in $Y$ which is connected to all nodes in the combination) through the matching, and therefore we necessarily have $q \leq {p \choose n}$.
\end{remark}

Finally, note that the existence of perfect $n$-gram matching results the existence of perfect $(n+1)$-gram matching\,\footnote{Note that the degree of each node (on matching side $Y$) in the original bipartite graph should be at least $n+1$.}, but the reverse is not true. For example, the bipartite graph $G(Y,X;E)$ with $|X|=4$ and $|Y|={4 \choose 2}=6$ in Figure \ref{fig:perfect ngram example}, has a perfect 2-gram matching, but not a perfect (1-gram) matching (since $6>4$). 

\subsubsection*{Identifiability conditions based on existence of perfect $n$-gram matching in topic-word graph}
Now, we are ready to propose the identifiability conditions and result. 

\begin{condition}[Perfect $n$-gram matching on $A$] \label{cond:perfect ngram matching}
The bipartite graph $G(V_h,V_o;A)$  between hidden and observed variables, has a perfect $n$-gram matching.
\end{condition}

The above condition implies that the sparsity pattern of matrix $A$ is appropriately scattered in the mapping from hidden to observed variables to be identifiable. Intuitively, it means that every hidden node can be distinguished from another hidden node by its unique set of neighbors under the corresponding $n$-gram matching. \\
Furthermore, condition \ref{cond:perfect ngram matching} is the key to be able to propose identifiability in the overcomplete regime. As stated in the size bound in Remark \ref{remark:ngram perfect matching necessary bound}, for $n \geq 2$, the number of hidden variables can be more than the number of observed variables and we can still have perfect $n$-gram matching. 

\begin{definition}[Kruskal rank, \cite{Kruskal:77}] The \textnormal{Kruskal rank} or the \textnormal{krank} of matrix $A$ is defined as the maximum number $k$ such that every subset of $k$ columns of $A$ is linearly independent.
\end{definition}

Note that krank is different from the general notion of matrix rank and it is a lower bound for the matrix rank, i.e., $\rank(A) \geq \krank(A)$.

\begin{condition}[Krank condition on $A$] \label{cond:krank bound}
The Kruskal rank of matrix $A$ satisfies the bound $\krank(A) \geq d_{\max}(A)^n$, where $d_{\max}(A)$ is the maximum node degree of any column of $A$.
\end{condition}

In the overcomplete regime, it is not possible for $A$ to be full column rank and $\krank(A) < |V_h| =q$. However, note that a large enough krank ensures that appropriate sized subsets of columns of $A$ are linearly independent. For instance, when $\krank(A)>1$, any two columns cannot be collinear and the above condition rules out the collinear case for identifiability. In the above condition, we see that a larger krank can incorporate denser connections between topics and words. 


The main identifiability result under a fixed graph structure is stated in the following theorem for $n \geq 2$, where $n$ is the topic persistence level. The identifiability result relies on having access to the $(2rn)$-th order moment of observed variables $x_l, l \in [2rn]$, defined in equation \eqref{eq:4th order moment of x definition} as
\begin{align*}
M_{2rn}(x) := \Ebb \left[ (x_1 \otimes x_2 \otimes \dotsm \otimes x_{rn})(x_{rn+1} \otimes x_{rn+2} \otimes \dotsm \otimes x_{2rn})^\top \right] \in \Rbb^{p^{rn} \times p^{rn}},
\end{align*}
for some integer $r \geq 1$.
\begin{theorem}[Generic identifiability under deterministic topic-word graph structure] \label{Thm:Identifiability based on A}
Let $M^{(n)}_{2rn}(x)$ in equation \eqref{eq:4th order moment of x definition} be the $(2rn)$-th order observed moment of the $n$-persistent topic model for some integer $r \geq 1$. If the model satisfies conditions \ref{cond:non-degeneracy}, \ref{cond:perfect ngram matching} and \ref{cond:krank bound}, then, for any $n \geq 2$, all the columns of population structure $A$ are \textnormal{generically identifiable} from $M^{(n)}_{2rn}(x)$. Furthermore, the $(2r)$-th order moment of the hidden variables, denoted by $M_{2r}(h)$, is also \textnormal{generically identifiable}.
\end{theorem}
The theorem is proved in Appendix \ref{Appendix:ProofThm_Identifiability based on A}.
It is seen that the population structure $A$ is identifiable, given any observed moment of order at least $2n$. Increasing the order of observed moment results in identifying higher order moments of the hidden variables. \\
The above theorem does not cover the case when the persistence level $n=1$. This is the usual bag-of-words admixture model. Identifiability of this model has been studied earlier \cite{AnandkumarEtal:LinearBayesianLatent} and we recall it below.

\begin{remark}[Bag-of-words admixture model, \cite{AnandkumarEtal:LinearBayesianLatent}] \label{remark:n=1 ident. result}
Given $(2r)$-th order observed moments with $r \geq 1$, the structure of the popular bag-of-words admixture model and the $(2r)$-th order moment of hidden variables are identifiable, when $A$ is full column rank and the following expansion condition holds
\cite{AnandkumarEtal:LinearBayesianLatent}
\begin{align} \label{eq:expansion n=1}
|N_A(S)| \geq |S| + d_{\max}(A), \quad \forall S \subseteq V_h, \ |S| \geq 2.
\end{align}
Our result for $n \geq 2$ in Theorem \ref{Thm:Identifiability based on A}, provides identifiability in the overcomplete regime with weaker matching condition \ref{cond:perfect ngram matching} and krank condition \ref{cond:krank bound}. The matching condition \ref{cond:perfect ngram matching} is weaker than the above expansion condition which is based on the perfect matching and hence, does not allow overcomplete models. Furthermore, the above result for the bag-of-words admixture model requires full column rank of $A$ which is  more stringent than our krank condition \ref{cond:krank bound}.
\end{remark}


\begin{remark}[Kruskal rank and degree diversity]
Condition~\ref{cond:krank bound} requires that the Kruskal rank of the topic-word matrix be large enough compared to the maximum degree of the topics. Intuitively, a larger Kruskal rank ensures enough diversity in the word supports among different topics under a higher level of sparsity. This Kruskal rank condition also allows for more degree diversity among the topics, when the topic persistence level $n>1$. On the other hand, for the bag-of-words model ($n=1$), using  \eqref{eq:expansion n=1} implies that $2d_{\min} > d_{\max}$, where $d_{\min}, d_{\max}$ are the minimum and maximum degrees of the topics. Thus, we provide identifiability results with more degree diversity when higher order moments are employed.
\end{remark}

\begin{remark}[Recovery using $\ell_1$ optimization]It turns out that our conditions for identifiability imply that the columns of the $n$-gram matrix $A^\ngram$, defined in Definition \ref{def:ngram matrix}, are the sparsest vectors in $\Col\Bigl( M^{(n)}_{2n}(x) \Bigr)$, having a tensor rank of one. See Appendix \ref{Appendix:ProofThm_Identifiability based on A}. 
This implies recovery of the columns of $A$ through   exhaustive search, which is not efficient. Efficient   $\ell_1$-based recovery algorithms have been analyzed  in \cite{SpielmanEtal2012,AnandkumarEtal:LinearBayesianLatent} for the undercomplete case  $(n=1)$. They can be employed here for recovery from higher order moments as well. Exploiting additional structure present in $A^{\ngram}$, for $n >1$, such as rank-1 test devices proposed in~\cite{DeLathauwerEtal:FOOBI} are interesting avenues for future investigation.
\end{remark}

\subsection{Analysis under random topic-word graph structures}  \label{sec:ident. conditions random}
In this section, we specialize the identifiability result to the random case. This result is based on more transparent conditions on the size and the degree of the random bipartite graph $G(V_h,V_o;A)$. We consider the random model where in the bipartite graph $G(V_h,V_o;A)$, each node $i \in V_h$ is randomly connected to $d_i$ different nodes in set $V_o$. Note that this is a heterogeneous degree model.

\begin{condition}[Size condition] \label{cond: size cond}
The random bipartite graph $G(V_h,V_o;A)$ with $|V_h|=q, |V_o|=p$, and $A \in \Rbb^{p \times q}$, satisfies the size condition $q \leq \bigl( c \frac{p}{n} \bigr)^n$ for some constant $0<c<1$. 
\end{condition}
This size condition is required to establish that the random bipartite graph has a perfect $n$-gram matching (and hence satisfies deterministic condition \ref{cond:perfect ngram matching}). It is shown in Section \ref{sec:perfect n-gram matching random} that the necessary size constraint $q=O(p^n)$ stated in Remark \ref{remark:ngram perfect matching necessary bound}, is achieved in the random case. Thus, the above constraint allows for the overcomplete regime,  where $q \gg p$ for $n \geq 2$, and is tight.
\begin{condition}[Degree condition] \label{cond: degree cond}
In the random bipartite graph $G(V_h,V_o;A)$ with $|V_h|=q, |V_o|=p$, and $A \in \Rbb^{p \times q}$, the degree $d_i$ of nodes $i \in V_h$ satisfies the following lower and upper bounds $(d_i \in [d_{\min}, d_{\max}])$:
\begin{itemize}
\item \textnormal{Lower bound:} $d_{\min} \geq \max \{ 1+\beta \log p, \alpha \log p \}$ for some constants $\beta > \frac{n-1}{\log 1/c}, \alpha > \max \bigl\{ 2n^2 \bigl( \beta \log \frac{1}{c} +1  \bigr), 2 \beta n \bigr\}$.
\item \textnormal{Upper bound:} $d_{\max} \leq (cp)^{\frac{1}{n}}$.
\end{itemize}
\end{condition}
Intuitively, the lower bound on the degree is required to show that the corresponding bipartite graph $G(V_h,V_o;A)$ has sufficient number of random edges to ensure that it has perfect $n$-gram matching with high probability. The upper bound on the degree is mainly required to satisfy the krank condition \ref{cond:krank bound}, where $d_{\max}(A)^n \leq \krank(A)$.

It is important to see that, for $n \geq 2$, the above condition on degree covers a range of models from sparse to intermediate regimes and it is reasonable in a number of applications that each topic does not generate a very large number of words.

\begin{definition}[\textbf{whp}]
A sequence of events $\mathcal{E}_p$ occurs \textnormal{with high probability} (\textbf{whp}) if $\Pr(\mathcal{E}_p) = 1 - O(p^{-\epsilon})$ for some $\epsilon>0$.
\end{definition}

The main random identifiability result is stated in the following theorem for $n \geq 2$, while $n=1$ case is addressed in Remark \ref{remark:n=1 ident. result random}.
The identifiability result relies on having access to the $(2rn)$-th order moment of observed variables $x_l, l \in [2rn]$, defined in equation \eqref{eq:4th order moment of x definition} as
\begin{align*}
M_{2rn}(x) := \Ebb \left[ (x_1 \otimes x_2 \otimes \dotsm \otimes x_{rn})(x_{rn+1} \otimes x_{rn+2} \otimes \dotsm \otimes x_{2rn})^\top \right] \in \Rbb^{p^{rn} \times p^{rn}},
\end{align*}
for some integer $r \geq 1$.

\textbf{Probability rate constants:} The probability rate of success in the following random identifiability result  is specified by constants $\beta' > 0$ and $\gamma = \gamma_1 + \gamma_2 > 0$ as
\begin{align}
\beta' & = -\beta  \log c - n+1, \label{eq:probability rate constants_beta'} \\
\gamma_1 & = e^{n-1} \Bigl( \frac{c}{n^{n-1}} + \frac{e^2}{1-\delta_1} n^{\beta'+1} \Bigr), \label{eq:probability rate constants_gamma1} \\
\gamma_2 & = \frac{c^{n-1} e^2}{n^n(1 - \delta_2)}, \label{eq:probability rate constants_gamma2}
\end{align}
where $\delta_1$ and $\delta_2$ are some constants satisfying $e^2 \Bigl( \frac{p}{n} \Bigr)^{- \beta \log 1/c} < \delta_1 <1$ and $\frac{c^{n-1} e^2}{n^n} p^{-\beta'} < \delta_2 <1$.

\begin{theorem}[Random identifiability] \label{Thm:Identifiability random}
Let $M^{(n)}_{2rn}(x)$ in equation \eqref{eq:4th order moment of x definition} be the $(2rn)$-th order observed moment of the $n$-persistent topic model for some integer $r \geq 1$. If the model with random population structure $A$ satisfies conditions \ref{cond:non-degeneracy}, \ref{cond: size cond} and \ref{cond: degree cond}, then \textbf{whp} (with probability at least $1-\gamma p^{-\beta'}$ for constants $\beta' > 0$ and $\gamma > 0$, specified in \eqref{eq:probability rate constants_beta'}-\eqref{eq:probability rate constants_gamma2}), for any $n \geq 2$, all the columns of population structure $A$ are identifiable from $M^{(n)}_{2rn}(x)$. Furthermore, the $(2r)$-th order moment of hidden variables, denoted by $M_{2r}(h)$, is also identifiable, \textbf{whp}.

\end{theorem}

The theorem is proved in Appendix \ref{Appendix:ProofThm_Identifiability random}. Similar to the deterministic analysis, it is seen that the population structure $A$ is identifiable given any observed moment with order at least $2n$. Increasing the order of observed moment results in identifying higher order moments of the hidden variables.\\

\begin{remark}[Trade-off between topic-word size ratio and degree]When the number of hidden variables increases, i.e. $c$ increases, but the order $n$ is kept fixed, the bounds on degree in condition~\ref{cond: degree cond} also needs to grow. Intuitively, a larger degree is needed to provide more flexibility in choosing the subsets of neighbors for hidden nodes to ensure the existence of a perfect $n$-gram matching in the bipartite graph, which in turn ensures identifiability. Note that as $c$ grows, the parameter $\beta$, which is the lower bound on $d$ also grows, and the probability rate (i.e., the term $-\beta \log c$) remains constant. Hence, the probability rate does not change as $c$ increases, since the increase in the degree $d$ compensates the additional ``difficulty'' arising  due to a larger number of hidden variables.
\end{remark}
The above identifiability theorem only covers for $n \geq 2$ and the $n=1$ case is addressed in the following remark.
\begin{remark}[Bag-of-words admixture model] \label{remark:n=1 ident. result random}
The identifiability result for the random bag-of-words admixture model is comparable to the result in  \cite{SpielmanEtal2012}, which considers exact recovery of sparsely-used dictionaries. They assume that $Y=DX$ is given for some unknown arbitrary dictionary $D\in \Rbb^{q\times q}$ and unknown random sparse coefficient matrix $X\in\Rbb^{q\times p}$. They establish that if $D\in \Rbb^{q\times q}$ is full rank and the random sparse coefficient matrix $X\in \Rbb^{q\times p}$ follows the Bernoulli-subgaussian model with size constraint $p > C q \log q$ and degree constraint $ O(\log q) < \Ebb[d] < O(q \log q)$, then the model is identifiable, whp. Comparing the size and degree constraints, our identifiability result for $n \geq 2$ requires more stringent upper bound on the degree ($d=O(p^{1/n})$), while more relaxed condition on the size ($q=O(p^n)$) which allows to identifiability in the overcomplete regime.
\end{remark}

\begin{remark}[The size condition is tight] \label{remark:size bound necessity and sufficiency}
The  size bound $q=O(p^n)$ in the above theorem achieves the necessary condition that $q \leq {p \choose n} = O(p^n)$ (see Remark~\ref{remark:ngram perfect matching necessary bound}), and is therefore tight. The sufficiency is argued in Theorem~\ref{Thm:perfect n-gram matching random graph}, where we show that the matching condition \ref{cond:perfect ngram matching} holds under the above size and degree conditions \ref{cond: size cond} and \ref{cond: degree cond}.
\end{remark}

\section{Identifiability  via Uniqueness of Tensor Decompositions} \label{sec:tensor decomp.}
In this section, we characterize the moments of the $n$-persistent topic model   in terms of the model parameters, i.e. the topic-word matrix $A$ and the moment of hidden variables.  We relate identifiability of the topic model   to uniqueness of a certain class of tensor decompositions, which in turn, enables us to prove Theorems \ref{Thm:Identifiability based on A} and \ref{Thm:Identifiability random}. We then discuss the special cases of the persistent topic model, \viz the single topic model (infinite-persistent topic model) and the bag-of-words admixture model (1-persistent topic model).

\subsection{Moment characterization of the persistent topic model} \label{sec:persistent topic model moment characterization}
The moment characterization requires the following definition of a $n$-gram matrix.
\begin{definition}[$n$-gram Matrix] \label{def:ngram matrix}
Given a matrix $A \in \R^{p \times q}$, its $n$-gram matrix $A^\ngram \in \R^{p^n
\times q}$ is defined as the matrix whose $(\bfi,j)$-th entry is given
by, for $\bfi:=(i_1,i_2,\dotsc,i_n)\in [p]^n$ and $j \in [q]$,
\begin{align*}
A^\ngram(\bfi, j)
:= A_{i_1,j} A_{i_2,j} \dotsb A_{i_n,j}, \quad \text{or} \quad A^\ngram:=\overbrace{A \odot \dotsm \odot A}^{n \operatorname{times}}.
\end{align*}
\end{definition}
That is, $A^\ngram$ is the column-wise $n^{\tha}$ order Kronecker product of $n$ copies of $A$, and is known as the  Khatri-Rao product~\cite{Golub&VanLoan:book}.

In the following lemma, which is proved in Appendix \ref{Appendix:proofoflem_persistent topic model moment characterization}, we characterize the observed moments of a persistent topic model. Throughout this section, the order of the observed moment is fixed to $2m$.


\begin{lem}[$n$-persistent topic model moment characterization] \label{lem:persistent topic model moment characterization}
The $(2m)$-th order moment of observed variables, defined in equation \eqref{eq:4th order moment of x definition}, for the $n$-persistent topic model is characterized as\,\footnote{The other cases not covered in Lemma~\ref{lem:persistent topic model moment characterization} are deferred to Appendix~\ref{Appendix:proofoflem_persistent topic model moment characterization}. See Remark~\ref{remark:moment lemma more derivaition}.}:
\begin{itemize}
\item  if $m=rn$, for some integer $r \geq 1$, then
\begin{align} \label{eq:2m-th order moment of x in terms of h}
M^{(n)}_{2m}(x) = \biggl( \overbrace{A^\ngram  \otimes \dots \otimes A^\ngram}^{r \operatorname{\ times}} \biggr) M_{2r}(h) \biggl( \overbrace{A^\ngram  \otimes \dots \otimes A^\ngram}^{r \operatorname{\ times}} \biggr)^\top,
\end{align}
where  $M_{2r}(h) \in \Rbb^{q^r \times q^r}$ is the $(2r)$-th order moment of hidden variables $h \in \Rbb^q$, defined in equation \eqref{eq:4th order moment of h}.
\item If $n \geq 2m$, then
\begin{align} \label{eq:4th order moment of x in terms of h_2ngram}
M^{(n)}_{2m}(x) = \left( A^\mgram\right) M_1(h)  \left( A^\mgram\right)^\top,
\end{align}
where $M_1(h) := \Diag(\Ebb[h]) \in \Rbb^{q \times q}$ is the first order moment of hidden variables $h \in \Rbb^q$, stacked in a diagonal matrix.
\end{itemize}
\end{lem}

Thus, we see that the observed moments can be expressed in terms of the hidden moments $M(h)$ and the Kronecker products of the $n$-gram matrices. In the special case, when the persistence level is large enough compared to the order of the moment $(n \geq 2m)$, the moment form reduces to a Khatri-Rao product form in \eqref{eq:4th order moment of x in terms of h_2ngram}. Moreover, in \eqref{eq:4th order moment of x in terms of h_2ngram}, we have a diagonal matrix $M_1(h)$ instead of a general (dense) matrix $M_{2r}(h)$ in \eqref{eq:2m-th order moment of x in terms of h},  when $n<2m= 2rn$. Thus, we have a more succinct representation of the moments in \eqref{eq:4th order moment of x in terms of h_2ngram} when the persistence level of the topics is large enough.

\begin{figure} \centering
\begin{minipage}[b]{2.2in}
\bp\psfrag{h}[l]{\scriptsize$h$}
\psfrag{y}[l]{\scriptsize$y$}
\psfrag{x1}[l]{\scriptsize$x_1$}\psfrag{xn}[l]{\scriptsize$x_m$}
\psfrag{xn1}[l]{\scriptsize$x_{m\!+\!1}$}\psfrag{xnn}[l]{\scriptsize$x_{2m}$}
\centering \includegraphics[width=1.8in]{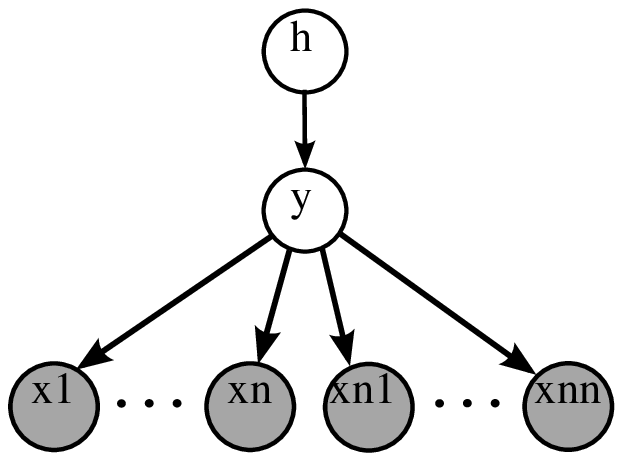}\ep
\subcaption{\small Single topic model \\ (infinite-persistent topic model)} \label{fig:MultiViewModel2n}
\end{minipage}
\hfil
\begin{minipage}[b]{2.2in}
\bp\psfrag{h}[l]{\scriptsize$h$}
\psfrag{y1}[l]{\scriptsize$y_1$}\psfrag{yn}[l]{\scriptsize$y_m$}
\psfrag{yn1}[l]{\scriptsize$y_{m\!+\!1}$}\psfrag{ynn}[l]{\scriptsize$y_{2m}$}
\psfrag{x1}[l]{\scriptsize$x_1$}\psfrag{xn}[l]{\scriptsize$x_m$}
\psfrag{xn1}[l]{\scriptsize$x_{m\!+\!1}$}\psfrag{xnn}[l]{\scriptsize$x_{2m}$}
\centering \includegraphics[width=1.8in]{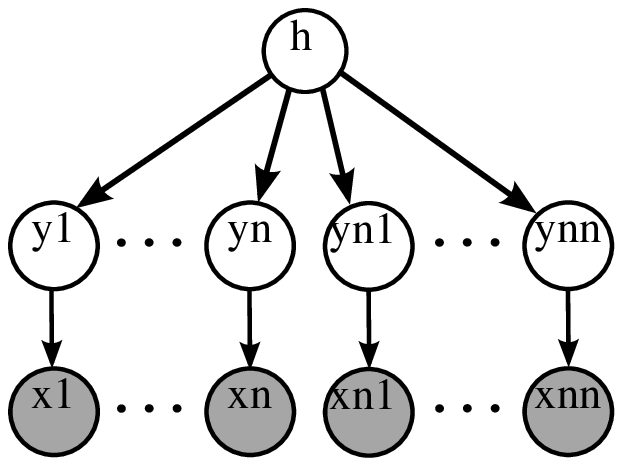}\ep
\subcaption{\small Bag-of-words admixture model \\ (1-persistent topic model)} \label{fig:MultiViewModelRegular}
\end{minipage}
\caption{\small
Hierarchical structure of the single topic model and bag-of-words admixture model shown for $2m$ number of words (views).}
\label{fig:MultiViewModel}
\end{figure}

In the following, we contrast the special cases when the persistence level $n$ is $n\to \infty$ (single topic model) and $n =1$ (bag of words admixture model), as shown in Fig.\ref{fig:MultiViewModel2n} and Fig.\ref{fig:MultiViewModelRegular}. In order to have a fair comparison, the number of observed variables is fixed to $2m$ and the persistence level is varied.

\textbf{Single topic model ($n \rightarrow \infty$): }The condition in \eqref{eq:4th order moment of x in terms of h_2ngram} $(n \geq 2m)$ is always satisfied for the single-topic  model, since $n \to \infty$ in this case, and we have
\begin{align} \label{eq:4th order moment of x in terms of h_2ngram-2}
M^{(\infty)}_{2m}(x) = \left( A^\mgram\right) M_1(h)  \left( A^\mgram\right)^\top.
\end{align} Note that $M_1(h)$ is a diagonal matrix.


\textbf{Bag-of-words admixture model ($n=1$): } From Lemma \ref{lem:persistent topic model moment characterization}, the $(2m)$-th order moment of observed variables $x_l, l \in [2m]$, for the bag-of-words admixture model (1-persistent topic model), shown in Figure \ref{fig:MultiViewModelRegular}, is given by
\begin{align} \label{eq:4th order moment of x in terms of h_regular}
M^{(1)}_{2m}(x) = \Bigl( \overbrace{A \otimes \dotsm \otimes A}^{m \operatorname{times}} \Bigr) M_{2m}(h)  \Bigl( \overbrace{A \otimes \dotsm \otimes A}^{m \operatorname{times}} \Bigr)^\top,
\end{align}
where $M_{2m}(h) \in \Rbb^{q^m \times q^m}$ is the $(2m)$-th order moment of hidden variables $h \in \Rbb^q$, defined in \eqref{eq:4th order moment of h}. Note that $M_{2m}(h)$ is a full matrix in general.

\textbf{Contrasting single topic $(n\to\infty)$ and bag of words models $(n=1)$: }Comparing equations \eqref{eq:4th order moment of x in terms of h_2ngram-2} and \eqref{eq:4th order moment of x in terms of h_regular}, it is seen that
the moments under the single topic model in \eqref{eq:4th order moment of x in terms of h_2ngram-2} are more ``structured'' compared to the bag of words model in \eqref{eq:4th order moment of x in terms of h_regular}. In \eqref{eq:4th order moment of x in terms of h_regular}, we have  Kronecker products of the topic-word matrix $A$, while  \eqref{eq:4th order moment of x in terms of h_2ngram-2}  involves   Khatri-Rao products of $A$. This forms a crucial criterion in determining of whether overcomplete models are identifiable, as discussed below.

\textbf{Why persistence helps in identifiability of overcomplete models? }For simplicity, let the order of the moment $2m=4$. The equations \eqref{eq:4th order moment of x in terms of h_2ngram-2} and \eqref{eq:4th order moment of x in terms of h_regular} reduce to
\begin{align}
M^{(\infty)}_{4}(x) & = (A \odot A) \Diag\left(\Ebb \bigl[ h]\right) (A \odot A)^\top, \label{eq:inf-persistent moment}\\
M^{(1)}_{4}(x) & = (A \otimes A) \Ebb \bigl[ (h \otimes h) (h \otimes h)^\top \bigr] (A \otimes A)^\top. \label{eq:1-persistent moment}
\end{align}
Note that for the single topic model in \eqref{eq:inf-persistent moment}, the   Khatri-Rao product  matrix $A \odot A \in \Rbb^{p^2 \times q}$ has  the same as the number of columns (i.e. the latent dimensionality) of the original matrix $A$, while the number of rows (i.e. the observed dimensionality) is increased. Thus,
the Khatri-Rao product ``expands'' the effect of hidden variables to higher order observed variables, which is the key towards identifying  overcomplete models. In other words, the original overcomplete representation becomes determined due to the `expansion effect' of the Khatri-Rao product structure of the higher order observed moments.

On the other hand, in the bag-of-words admixture model in \eqref{eq:1-persistent moment}, this interesting `expansion property' does not occur, and we have the   Kronecker product $A \otimes A \in \Rbb^{p^2 \times q^2}$, in place of the Khatri-Rao products. The Kronecker product operation increases both the number of the columns (i.e. latent dimensionality) and the number of rows (i.e. observed dimensionality), which implies that higher order moments do not help in identifying overcomplete models.

\begin{figure} \centering
\begin{minipage}[b]{6in} \centering
\bp
\psfrag{X}[l]{\scriptsize$X$}\psfrag{Y}[l]{\scriptsize$Y$}
\psfrag{1}[l]{\scriptsize$1$}\psfrag{2}[l]{\scriptsize$2$}\psfrag{3}[l]{\scriptsize$3$}\psfrag{4}[l]{\scriptsize$4$}\psfrag{5}[l]{\scriptsize$5$}
\includegraphics[width=2in]{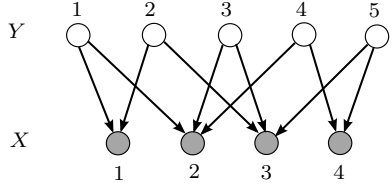}\ep
\vspace{0.1in}
\subcaption{\small Structure of an overcomplete matrix $A \in \Rbb^{4 \times 5}$ having a perfect 2-gram matching.}
\end{minipage}\\
\vspace{0.15in}
\begin{minipage}[b]{6in}
\bp
\psfrag{X2}[l]{\scriptsize$X^{(2)}$}\psfrag{Y}[l]{\scriptsize$Y$}
\psfrag{1}[l]{\scriptsize$1$}\psfrag{2}[l]{\scriptsize$2$}\psfrag{3}[l]{\scriptsize$3$}\psfrag{4}[l]{\scriptsize$4$}\psfrag{5}[l]{\scriptsize$5$}
\psfrag{11}[l]{\scriptsize$(1,1)$}
\psfrag{12}[l]{\scriptsize$(1,2)$}
\psfrag{13}[l]{\scriptsize$(1,3)$}
\psfrag{14}[l]{\scriptsize$(1,4)$}
\psfrag{21}[l]{\scriptsize$(2,1)$}
\psfrag{22}[l]{\scriptsize$(2,2)$}
\psfrag{23}[l]{\scriptsize$(2,3)$}
\psfrag{24}[l]{\scriptsize$(2,4)$}
\psfrag{31}[l]{\scriptsize$(3,1)$}
\psfrag{32}[l]{\scriptsize$(3,2)$}
\psfrag{33}[l]{\scriptsize$(3,3)$}
\psfrag{34}[l]{\scriptsize$(3,4)$}
\psfrag{41}[l]{\scriptsize$(4,1)$}
\psfrag{42}[l]{\scriptsize$(4,2)$}
\psfrag{43}[l]{\scriptsize$(4,3)$}
\psfrag{44}[l]{\scriptsize$(4,4)$}
\centering \includegraphics[width=6in]{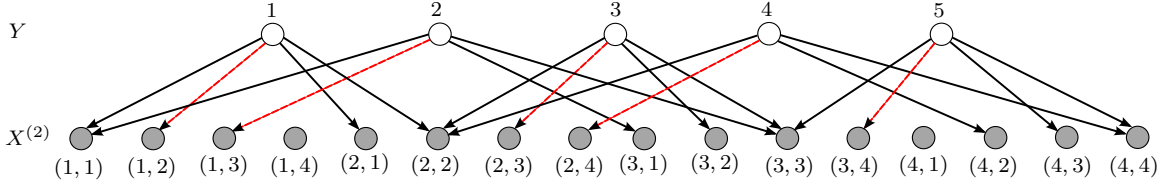}\ep
\vspace{0.1in}
\subcaption{\small Structure of $A \odot A \in \Rbb^{16 \times 5}$ having a perfect ($Y$-saturating) matching, highlighted by dashed red edges. 
}
\end{minipage} \\
\vspace{0.15in}
\begin{minipage}[b]{6in}
\bp
\psfrag{X2}[l]{\scriptsize$X^{(2)}$}\psfrag{Y2}[l]{\scriptsize$Y^{(2)}$}
\psfrag{11}[l]{\scriptsize$(1,1)$}
\psfrag{12}[l]{\scriptsize$(1,2)$}
\psfrag{13}[l]{\scriptsize$(1,3)$}
\psfrag{14}[l]{\scriptsize$(1,4)$}
\psfrag{15}[l]{\scriptsize$(1,5)$}
\psfrag{21}[l]{\scriptsize$(2,1)$}
\psfrag{22}[l]{\scriptsize$(2,2)$}
\psfrag{23}[l]{\scriptsize$(2,3)$}
\psfrag{24}[l]{\scriptsize$(2,4)$}
\psfrag{31}[l]{\scriptsize$(3,1)$}
\psfrag{32}[l]{\scriptsize$(3,2)$}
\psfrag{33}[l]{\scriptsize$(3,3)$}
\psfrag{34}[l]{\scriptsize$(3,4)$}
\psfrag{41}[l]{\scriptsize$(4,1)$}
\psfrag{42}[l]{\scriptsize$(4,2)$}
\psfrag{43}[l]{\scriptsize$(4,3)$}
\psfrag{44}[l]{\scriptsize$(4,4)$}
\psfrag{45}[l]{\scriptsize$(4,5)$}
\psfrag{51}[l]{\scriptsize$(5,1)$}
\psfrag{52}[l]{\scriptsize$(5,2)$}
\psfrag{53}[l]{\scriptsize$(5,3)$}
\psfrag{54}[l]{\scriptsize$(5,4)$}
\psfrag{55}[l]{\scriptsize$(5,5)$}
\centering \includegraphics[width=6in]{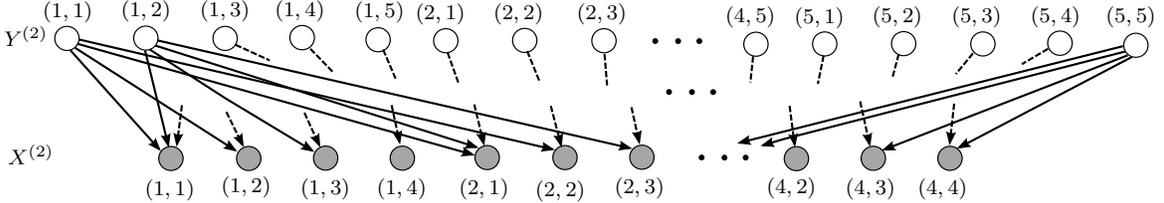}\ep
\vspace{0.1in}
\subcaption{\small Structure of $A \otimes A \in \Rbb^{16 \times 25}$. For simplicity, only a few edges and nodes are shown and the dashed edges denote the bunch of edges connected to each node, not specifically shown. 
}
\end{minipage}
\caption{\small An example of an overcomplete matrix $A$ and the matrices $A \odot A$ and $A \otimes A$. The corresponding bipartite graphs encode the sparsity pattern of each of the matrices. 
$A \odot A$ expands the effect of hidden variables to second order observed variables which is crucial for overcomplete identifiability, while in the $A \otimes A$, the order of both the hidden and observed variables are    increased.}
\label{fig:Kron_vs_KR}
\end{figure}

An example is provided in Figure \ref{fig:Kron_vs_KR} which helps to see how the matrices $A \odot A$ and $A \otimes A$ behave differently in terms of mapping topics to word tuples.

Note that for the $n$-persistent model, for $n=2$, the $4^{\tha}$ order moment reduces to
\beq
M^{(2)}_{4}(x)  = (A \odot A)  \Ebb \bigl[ h h^\top] (A \odot A)^\top. \label{eq:2-persistent moment}\eeq Contrasting the above equation with \eqref{eq:inf-persistent moment} and \eqref{eq:1-persistent moment}, we find that the $2$-persistent model retains the desirable property of possessing Khatri-Rao products, while being more general than the form for single topic model in \eqref{eq:inf-persistent moment}. This key property enables us to establish identifiability of topic models with finite persistence levels.

\subsection{Tensor algebra of the model} \label{sec:persistent topic model moment characterization tensor form}

In Section~\ref{sec:persistent topic model moment characterization}, we provided a representation of the moment forms in the matrix form. We now provide the equivalent tensor representation of the moments. The tensor representation is more compact and transparent, and allows us to compare the topic models under different levels of persistence.
We compare the derived tensor form with  the well-known Tucker and CP decompositions. We first introduce some tensor notations and definitions.


\subsubsection{Tensor notations and definitions} 
A real-valued order-$n$ tensor $A \in \bigotimes_{i=1}^n \Rbb^{p_i} := \Rbb^{p_1 \times \dotsm \times p_n}$ is a $n$ dimensional array $A(1:p_1,\dotsc,1:p_n)$, where the $i$-th mode is indexed from $1$ to $p_i$.  In this paper, we restrict ourselves to the case that $p_1 = \dotsb = p_n = p$, and simply write $A \in \bigotimes^n \Rbb^p$. A {\em  fiber} of a tensor $A$ is a vector obtained by fixing all indices of $A$ except one, e.g., for $A \in \bigotimes^4 \Rbb^3$, the vector $f = A(2,1:3,3,1)$ is a fiber. \ \\
For a vector $u \in \Rbb^p$, $\Diagn(u) \in \bigotimes^n \Rbb^p$ is the $n$-th order diagonal tensor with vector $u$ on its diagonal.
The tensor $A \in \bigotimes^n \Rbb^p$, is stacked as a vector $a \in \Rbb^{p^n}$ by the $\vecform(\cdot)$ operator, defined as
\begin{align*}
a = \vecform(A) \Leftrightarrow a \bigl( (i_1-1)p^{n-1} + (i_2-1)p^{n-2} + \dotsb + (i_{n-1}-1)p + i_n)  \bigr) = A(i_1,i_2,\dotsc,i_n).
\end{align*}
The inverse of $a = \vecform(A)$ operation is denoted by $A = \ten(a)$. \\
For vectors $a_i \in \Rbb^{p_i}, i \in [n]$, the tensor outer product operator ``$\circ$'' is defined as \cite{Golub&VanLoan:book}
\begin{align}
A = a_1 \circ a_2 \circ \dotsb \circ a_n \in \bigotimes_{i=1}^n \Rbb^{p_i} \Leftrightarrow A(i_1,i_2,\dotsc,i_n) := a_1(i_1) a_2(i_2) \dotsm a_n(i_n). \label{Def:tensor outer product}
\end{align}
The above generated tensor is a rank-1 tensor. The \textit{tensor rank} is the minimal number of rank-1 tensors into which a tensor can be decomposed. This type of rank is called CP (Candecomp/Parafac) tensor rank in the literature \cite{Golub&VanLoan:book}. \\
According to above definitions, for any set of vectors $a_i \in \Rbb^{p_i}, i \in [n]$, we have the following pair of equalities:
\begin{align*}
\vecform(a_1 \circ a_2 \circ \dotsb \circ a_n) = a_1 \otimes a_2 \otimes \dotsb \otimes a_n, \\
\ten(a_1 \otimes a_2 \otimes \dotsb \otimes a_n) = a_1 \circ a_2 \circ \dotsb \circ a_n.
\end{align*}
For any vector $a \in  \Rbb^p$, the power notations are also defined as
\begin{align*}
a^{\otimes n} := \overbrace{a \otimes a \otimes \dotsb \otimes a}^{n \operatorname{times}} \in \Rbb^{p^n}, \\
a^{\circ n} := \overbrace{a \circ a \circ \dotsb \circ a}^{n \operatorname{times}} \in \bigotimes^n \Rbb^p.
\end{align*}
The second power is usually called the $n$-th order \textit{tensor power} of vector $a$. \\
Finally, the Tucker and CP (Candecomp/Parafac) representations are defined as follows \cite{Golub&VanLoan:book,kolda_survey}.
\begin{definition}[Tucker representation] \label{def:Tucker decomp}
Given a core tensor $S \in \bigotimes_{i=1}^n \Rbb^{r_i}$ and inverse factors $U_i \in \Rbb^{p_i \times r_i}, i \in [n]$, the Tucker representation of the $n$-th order tensor $A \in \bigotimes_{i=1}^n \Rbb^{p_i}$ is
\begin{align} \label{eq:Tucker representation}
A = \sum_{i_1=1}^{r_1} \sum_{i_2=1}^{r_2} \dots \sum_{i_n=1}^{r_n} S(i_1,i_2,\dotsc,i_n) U_1(:,i_1) \circ U_2(:,i_2) \circ \dotsm \circ U_n(:,i_n) =: [[S; U_1, U_2, \dotsc, U_n]],
\end{align}
where $U_j(:,i_j)$ denotes the $i_j$-th column of matrix $U_j$. The tensor $S$ is referred to as the core tensor.
\end{definition}
\begin{definition}[CP representation] \label{def:CP decomp}
Given $\lambda \in \Rbb^r, U_i \in \Rbb^{p_i \times r}, i \in [n]$, the CP representation of the $n$-th order tensor $A \in \bigotimes_{i=1}^n \Rbb^{p_i}$ is
\begin{align} \label{eq:CP representation}
A = \sum_{i=1}^r \lambda_i U_1(:,i) \circ U_2(:,i) \circ \dotsm \circ U_n(:,i) =: [[\Diagn(\lambda); U_1, U_2, \dotsc, U_n]],
\end{align}
where $U_j(:,i)$ denotes the $i$-th column of matrix $U_j$. 
\end{definition}
Note that the CP representation is a special case of the Tucker representation when the core tensor $S$ is square and diagonal.

\subsubsection{Tensor representation of moments under topic model}

We now provide a tensor representation of the moments.

For the $n$-persistent topic model, the $2m$-th observed moment is denoted by $T^{(n)}_{2m}(x)$, which is the tensor form of the moment matrix $M^{(n)}_{2m}(x)$, characterized in Lemma \ref{lem:persistent topic model moment characterization}. It is given by
\begin{align} \label{eq:4th order moment of x definition tensor form}
T_{2m}(x)_{(i_1,i_2,\dotsc,i_{2m})} := \Ebb [x_1(i_1)x_2(i_2) \dotsm x_{2m}(i_{2m})], \quad i_1,i_2,\dotsc,i_{2m} \in [p],
\end{align}
where $T_{2m}(x) \in \bigotimes^{2m} \Rbb^p$.

This tensor is characterized in the following lemma, and is proved in Appendix \ref{Appendix:proofoflem_persistent topic model moment characterization}.

\begin{lem}[$n$-persistent topic model moment characterization in tensor form] \label{lem:persistent topic model moment characterization tensor form}
The $(2m)$-th order moment of words, defined in equation \eqref{eq:4th order moment of x definition tensor form}, for the $n$-persistent topic model is characterized as\,\footnote{The other cases not covered in Lemma~\ref{lem:persistent topic model moment characterization tensor form} are deferred to Appendix~\ref{Appendix:proofoflem_persistent topic model moment characterization}. See Remark~\ref{remark:moment lemma more derivaition}.}:
\begin{itemize}
\item  if $m=rn$ for some integer $r \geq 1$, then
\begin{align} \label{eq:4th order moment of x_tensor form}
T^{(n)}_{2m}(x)
& = \sum_{i_1=1}^{q} \sum_{i_2=1}^{q} \dots \sum_{i_{2r}=1}^{q} \Ebb[h_{i_1} h_{i_2} \dotsm h_{i_{2r}}] a_{i_1}^{\circ n} \circ a_{i_2}^{\circ n} \circ \dotsm \circ a_{i_{2r}}^{\circ n} \\
& = \Bigl[ \Bigl[ S_r; \overbrace{A,A,\dotsc,A}^{2m \ \operatorname{times}} \Bigr] \Bigr] \nn,
\end{align}
where $S_r \in \bigotimes^{2rn} \Rbb^q$ is the core tensor in  the above Tucker representation with the sparsity pattern as
\begin{align*}
S_r \bigl(\bfi \bigr) =
\left\{\begin{array}{ll}
M_{2r}(h)_{\bigl((i_n, i_{2n},\dotsc, i_{rn}),(i_{(r+1)n}, i_{(r+2)n},\dotsc, i_{2rn})\bigr)}  & ,i_1\!=\!i_2\!=\!\dotsb\!=\!i_n, i_{n+1}\!=\!i_{n+2}\!=\!\dotsb\!=\!i_{2n},\dotsc \\
0 & ,\operatorname{o.w.}, \nn
\end{array}\right.
\end{align*}
where $\bfi:=  (i_1,i_2,\dotsc,i_{2rn})$. 
\item If $n \geq 2m$, then
\begin{align} \label{eq:4th order moment of x in terms of h_2ngram tensor form}
T^{(n)}_{2m}(x) 
= \sum_{i \in [q]} \Ebb[h_i] a_i^{\circ 2m}
=\bigl[\bigl[ \Diag_{2m}(\Ebb[h]); \overbrace{A,A,\dotsc,A}^{2m \operatorname{\ times}} \bigr]\bigr].
\end{align}
\end{itemize}
\end{lem}

The tensor representation in \eqref{eq:4th order moment of x_tensor form} is a specific type of tensor decomposition which is  a special case of the Tucker representation (since $S_r$ is not fully dense), but more general than the CP representation. The tensor representation in \eqref{eq:4th order moment of x in terms of h_2ngram tensor form} has a CP form.


\subsubsection*{Comparison with single topic model and bag-of-words admixture model}
We now provide the tensor form for the special cases  single topic model and bag-of-words admixture model.
In order to have a fair comparison, the number of observed variables is fixed to $2m$ and the persistence level is varied. 

\textbf{CP representation of the single topic model: }
The $(2m)$-th order moment of the words for the single topic model (infinite-persistent topic model) is provided in equation \eqref{eq:4th order moment of x in terms of h_2ngram tensor form} as
\begin{align}
T^{(\infty)}_{2m}(x) 
= \sum_{i \in [q]} \Ebb[h_i] a_i^{\circ 2m} = \bigl[\bigl[ \Diag_{2m}(\Ebb[h]); \overbrace{A,A,\dotsc,A}^{2m \operatorname{\ times}} \bigr]\bigr]. \label{eqn:tensorninf}
\end{align}
This representation is the symmetric CP representation\,\footnote{In Appendix \ref{Appendix:CP uniqueness}, we provide a more detailed comparison between our approach and some of the previous identifiability results for the (overcomplete) CP decomposition.} of $T^{(\infty)}_{2m}(x)$.



\textbf{Tucker representation of the bag-of-words admixture model: }
From Lemma \ref{lem:persistent topic model moment characterization tensor form}, the tensor form of the $(2m)$-th order moment of observed variables $x_l, l \in [2m]$, for the bag-of-words admixture model (1-persistent topic model) is given by
\begin{align}
T^{(1)}_{2m}(x) 
& = \sum_{i_1=1}^{q} \sum_{i_2=1}^{q} \dots \sum_{i_{2m}=1}^{q} \Ebb[h_{i_1} h_{i_2} \dotsm h_{i_{2m}}] a_{i_1} \circ a_{i_2} \circ \dotsm \circ a_{i_{2m}} \nn\\
& = \Bigl[\Bigl[ \Ebb \bigl[ h^{\circ (2m)} \bigr]; \overbrace{A,A,\dotsc,A}^{2m \operatorname{\ times}} \Bigr]\Bigr].\label{eqn:tensorn1}
\end{align}
This representation is the Tucker representation (decomposition) of $T^{(1)}_{2m}(x)$ where the core tensor $S = \Ebb \bigl[ h^{\circ (2m)} \bigr]$ is the tensor form of the $(2m)$-th order hidden moment $M_{2m}(h)$, defined in equation \eqref{eq:4th order moment of h}, and the inverse factors   correspond to the population structure $A$. 

Comparing the tensor forms for the $n$-persistent topic model \eqref{eq:4th order moment of x_tensor form}, single topic model \eqref{eqn:tensorninf}, and bag of words admixture model \eqref{eqn:tensorn1}, we find that all of them involve Tucker decompositions, where the inverse factors correspond to the topic-word matrix $A$, and the only difference is in the sparsity level of the core tensor $S$. For the bag of words model, with $n=1$, the core tensor is fully dense in general, while for the single topic model, with $n \to\infty$, the core tensor is diagonal which reduces to the CP decomposition. For a general topic model with persistence level $n$, the core tensor is in between these two extremes and has structured sparsity. This  sparsity property of the core tensor is crucial towards  establishing identifiability in the overcomplete regime. The bag-of-words model is not identifiable in the overcomplete regime since the core tensor is fully dense in this case, while an overcomplete $n$-persistent topic model can be identified under certain constraints provided in Section \ref{sec:ident. conditions}, since the core tensor has structured sparsity and symmetry.



\section{Proof Techniques and Auxiliary Results}
The main identifiability results are given in Theorems \ref{Thm:Identifiability based on A} and \ref{Thm:Identifiability random} for deterministic and random cases of topic-word graph structures.
In this section, we provide a proof sketch of these results, and then, we propose auxiliary results on the existence of perfect $n$-gram matching for random bipartite graphs and a lower bound on the Kruskal rank of random matrices.

\subsection{Proof sketch} \label{sec:proof sketch}

\begin{figure}\centering \bp
\psfrag{SD}[l]{\scriptsize Size \& degree}
\psfrag{C45}[l]{\scriptsize conditions \ref{cond: size cond},\ref{cond: degree cond}}
\psfrag{FRC}[l]{\scriptsize for random case}
\psfrag{T34}[l]{\scriptsize Theorems \ref{Thm:perfect n-gram matching random graph},\ref{Thm:krank bound for random case}}
\psfrag{MK}[l]{\scriptsize Matching \& krank}
\psfrag{C23}[l]{\scriptsize conditions \ref{cond:perfect ngram matching},\ref{cond:krank bound}}
\psfrag{A}[l]{\scriptsize on $A$}
\psfrag{L5}[l]{\scriptsize Lemma \ref{lem:suff matching for rank and expansion}}
\psfrag{RE}[l]{\scriptsize Rank \& expansion}
\psfrag{C67}[l]{\scriptsize conditions \ref{cond:rank},\ref{cond:expansion}}
\psfrag{An}[l]{\scriptsize on $A^\ngram$}
\psfrag{T5}[l]{\scriptsize Theorem \ref{Thm:Identifiability}}
\psfrag{ND}[l]{\scriptsize Non-degeneracy}
\psfrag{C1}[l]{\scriptsize condition \ref{cond:non-degeneracy} on $h$}
\psfrag{IDEN}[l]{\scriptsize Identifiability}
\includegraphics[width=6.0in]{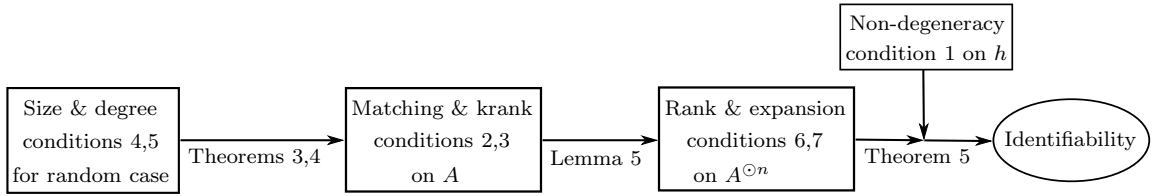} \ep
\caption{\small Hierarchy among the proposed conditions and results.} \label{fig:ProofHierarchy}
\end{figure}

\textbf{Summary of relationships among different conditions: }To summarize, there exists a hierarchy among the proposed conditions as follows. See Figure \ref{fig:ProofHierarchy}.
First, in the random analysis, the size and the degree conditions \ref{cond: size cond} and \ref{cond: degree cond} are sufficient for satisfying the perfect $n$-gram matching and the krank conditions \ref{cond:perfect ngram matching} and \ref{cond:krank bound}, shown by Theorems \ref{Thm:perfect n-gram matching random graph} and \ref{Thm:krank bound for random case}. Then, these conditions \ref{cond:perfect ngram matching} and \ref{cond:krank bound} ensure that the rank and the expansion conditions \ref{cond:rank} and \ref{cond:expansion} hold, shown by Lemma \ref{lem:suff matching for rank and expansion}.
And finally, these conditions \ref{cond:rank} and \ref{cond:expansion} together with non-degeneracy condition \ref{cond:non-degeneracy} conclude the primary identifiability result in Theorem \ref{Thm:Identifiability}. Note that the genericity of $A$ is also required for these results to hold.

\textbf{Primary deterministic analysis in Theorem \ref{Thm:Identifiability}: } The deterministic analysis is primarily based on conditions on the $n$-gram matrix $A^\ngram$; but since these conditions are opaque (mainly expansion condition on $A^\ngram$, provided in condition \ref{cond:expansion}), this analysis is related to conditions on matrix $A$ itself.
See Theorem \ref{Thm:Identifiability} in Appendix \ref{sec:ident. conditions based on A^ngram} for the identifiability result based on $A^\ngram$.
We briefly discuss it below for the case when $2n$ number of words are available under the $n$-persistent topic model.
From equation \eqref{eq:2m-th order moment of x in terms of h}, the $(2n)$-th order moment of the observed variables under the $n$-persistent topic model can be written as
\begin{align} \label{eq:4th order moment of x in terms of h}
M^{(n)}_{2n}(x) = \Bigl( A^\ngram \Bigr) \mathbb{E} \bigl[ h h^\top \bigr] \Bigl( A^\ngram \Bigr)^\top.
\end{align}
The question is whether we can recover $A$, given the $M^{(n)}_{2n}(x)$. Obviously, the matrix $A$ is not identifiable without any further conditions.
First, non-degeneracy and rank conditions (conditions \ref{cond:non-degeneracy} and \ref{cond:rank}) are required. 
Assuming these two conditions, we have from \eqref{eq:4th order moment of x in terms of h} that
\[
\Col \Bigl( M^{(n)}_{2n}(x) \Bigr) = \Col \Bigl( A^\ngram \Bigr).
\]
Therefore, the problem of recovering $A$ from $M^{(n)}_{2n}(x)$ reduces to finding $A^\ngram$ in $\Col \bigl( A^\ngram \bigr)$. \\
Then, we show that under the following expansion condition on $A^\ngram$ and the genericity property, matrix $A$ is identifiable from $\Col \bigl( A^\ngram \bigr)$. The expansion condition (refer to condition \ref{cond:expansion} for a more detailed statement), imposes the following property on the bipartite graph $G \bigl( V_h,V_o^{(n)};A^\ngram \bigr)$\,\footnote{$V_o^{(n)}$ denotes all ordered $n$-tuples generated from set $V_o := \{1,\dotsc,p\}$ which indexes the rows of $A^{\ngram}$.},
\begin{align} \label{eq:expansion}
\left| N_{A_\Rest^\ngram}(S) \right| \geq |S| + d_{\max} \Bigl( A^\ngram \Bigr), \quad \forall S \subseteq V_h, \ |S| > \krank(A),
\end{align}
where $d_{\max} \bigl( A^\ngram \bigr)$ is the maximum node degree in set $V_h$, and the restricted version of $n$-gram matrix, denoted by $A_\Rest^\ngram$, is obtained by removing its redundant (identical) rows (see Definition \ref{def:ngram matrix restricted}).
The identifiability claim is proved by showing that the columns of $A^\ngram$ are the sparsest and rank-1 vectors (in the tensor form) in $\Col\bigl( A^\ngram \bigr)$ under the expansion condition in \eqref{eq:expansion} and genericity conditions. 
Note that since we only require expansion on sets larger than Kruskal rank, the expansion condition \eqref{eq:expansion} is a more relaxed condition compared to expansion condition proposed in \cite{AnandkumarEtal:LinearBayesianLatent,SpielmanEtal2012} for identifiability in the undercomplete regime. For a more detailed comparison, refer to Remark \ref{remark:conditions modifications} in Appendix \ref{sec:ident. conditions based on A^ngram}.

\textbf{Deterministic analysis in Theorem \ref{Thm:Identifiability based on A}: }Expansion and rank conditions in Theorem \ref{Thm:Identifiability} are imposed on the $n$-gram matrix $A^\ngram$. According to the generalized matching notions, defined in Section \ref{sec:ident. conditions deterministic}, sufficient combinatorial conditions on matrix $A$ (conditions \ref{cond:perfect ngram matching} and \ref{cond:krank bound}) are introduced which ensure that the expansion and rank conditions on $A^\ngram$ are satisfied. 
The following lemma is employed to establish these results, where we state an interesting property which relates the existence of a  perfect matching in $A^\ngram$ to the existence of a  perfect $n$-gram matching in $A$.
\begin{lem} \label{lem:ngram perfect matching property}
If $G(Y,X;A)$ has a perfect $n$-gram matching, then $G(Y,X^{(n)};A^{\ngram})$ has a perfect matching. In the other direction, if $G(Y,X^{(n)};A^{\ngram})$ has a perfect matching $M^{\ngram}$, then $G(Y,X;A)$ has a perfect $n$-gram matching under the following condition on $M^{\ngram}$. All the matching edges $(j,(i_1,\dotsc,i_n)) \in M^{\ngram}$ should satisfy $i_1 \neq i_2 \neq \dotsb \neq i_n$ for all $j \in Y$. In words, the matching edges should be connected to nodes in $X^{(n)}$, which are indexed by tuples of distinct indices.
\end{lem}
See Appendix \ref{Appendix:AuxLemma} for a proof.
Using this lemma, condition \ref{cond:perfect ngram matching} implies that $G(Y,X^{(n)};A^{\ngram})$ has a perfect matching. Then, it is straightforward to argue that the expansion and rank conditions on $A^\ngram$ are satisfied, which is shown in Lemma \ref{lem:suff matching for rank and expansion} in Appendix \ref{Appendix:lem_suff matching for rank and expansion}.
This leads to the generic identifiability result stated in Theorem \ref{Thm:Identifiability based on A}.

\subsection{Analysis of Random Structures} \label{sec:random proof sketch}
The identifiability result for a random structured matrix $A$ is provided in Theorem \ref{Thm:Identifiability random}. Sufficient size and degree conditions \ref{cond: size cond} and \ref{cond: degree cond} on the random matrix $A$ are proposed such that the deterministic combinatorial conditions \ref{cond:perfect ngram matching} and \ref{cond:krank bound} on $A$ are satisfied. The details of these auxiliary results are provided in the following two subsequent sections.
In Section \ref{sec:perfect n-gram matching random}, it is proved in Theorem \ref{Thm:perfect n-gram matching random graph} that a random bipartite graph satisfying reasonable size and degree constraints, has a perfect $n$-gram matching (condition \ref{cond:perfect ngram matching}), \textbf{whp}.
Then, a lower bound on the Kruskal rank of a random matrix $A$ under size and degree constraints is provided in Theorem \ref{Thm:krank bound for random case} in Section \ref{sec:Lower bound on the Kruskal rank random}, which implies the krank condition \ref{cond:krank bound}. Intuitions on why such size and degree conditions are required, are mentioned in Section \ref{sec:ident. conditions random} where these conditions are proposed.

\subsubsection{Existence of perfect $n$-gram matching for random bipartite graphs} \label{sec:perfect n-gram matching random}
We show in the following theorem that a random bipartite graph satisfying reasonable size and degree constraints, proposed earlier in conditions \ref{cond: size cond} and \ref{cond: degree cond}, has a perfect $n$-gram matching \textbf{whp}.

\begin{theorem}[Existence of perfect $n$-gram matching for random bipartite graphs] \label{Thm:perfect n-gram matching random graph}
Consider a random bipartite graph $G(Y,X;E)$ with $|Y|=q$ nodes on the left side and $|X|=p$ nodes on the right side, and each node $i \in Y$ is randomly connected to $d_i$ different nodes in $X$. Let $d_{\min} := \min_{i \in Y} d_i$. Assume that it satisfies the size condition $q \leq \bigl( c \frac{p}{n} \bigr)^n$ (condition \ref{cond: size cond}) for some constant $0 < c <1$  and the degree condition $d_{\min} \geq \max \{1+\beta \log p, \alpha \log p \}$ for some constants $\beta > \frac{n-1}{\log 1/c}, \alpha > \max \bigl\{ 2n^2 \bigl( \beta \log \frac{1}{c} +1  \bigr), 2 \beta n \bigr\}$ (lower bound in condition \ref{cond: degree cond}).
Then, there exists a perfect ($Y$-saturating) $n$-gram matching in the random bipartite graph $G(Y,X;E)$, with probability at least $1-\gamma_1 p^{-\beta'}$ for constants $\beta' > 0$ and $\gamma_1 > 0$, specified in \eqref{eq:probability rate constants_beta'} and \eqref{eq:probability rate constants_gamma1}.
\end{theorem}



Note that the sufficient size bound $q=O(p^n)$ in the above theorem is also necessary (see Remark \ref{remark:ngram perfect matching necessary bound}), and is therefore tight.

\begin{remark}[Insufficiency of the union bound argument] \label{remark:Insuff. of union bound}
It is easier to exploit the union bound  arguments to propose random bipartite graphs which have a perfect $n$-gram matching \textbf{whp}. 
It is proved in Appendix \ref{Appendix:ProofThm_Auxiliary Random Theorems} that if $d \geq n$ and the size constraint $|Y| = O(|X|^{\frac{n}{2}-\delta})$ for some $\delta > 0$ is satisfied, then \textbf{whp}, the random bipartite graph has a perfect $n$-gram matching.
Comparing this result with ours in Theorem \ref{Thm:perfect n-gram matching random graph}, our approach has a better size scaling while the union bound approach has a better degree scaling. The size scaling limitation in the union bound argument makes it unattractive. In order to identify the population structure $A$ in the overcomplete regime where $|Y| = O(|X|^n)$, we need access to at least $(4n)$-th order moment under the union bound argument, while only the $(2n)$-th order moment is required under our argument. 
\end{remark}

\subsubsection{Lower bound on the Kruskal rank of random matrices} \label{sec:Lower bound on the Kruskal rank random}
In the following theorem, a lower bound on the Kruskal rank of a random matrix $A$ under dimension and degree constraints is provided, which is proved in Appendix \ref{Appendix:ProofThm_Auxiliary Random Theorems}. \\

\begin{theorem}[Lower bound on the Kruskal rank of random matrices] \label{Thm:krank bound for random case}
Consider a random matrix $A \in \Rbb^{p \times q}$, where for any $i \in [q]$, there are $d_i$ number of random non-zero entries in column $i$. Let $d_{\min} := \min_{i \in [q]} d_i$.  Assume that it satisfies the size condition $q \leq \bigl( c \frac{p}{n} \bigr)^n$ (condition \ref{cond: size cond}) for some constant $0<c<1$ and the degree condition $d_{\min} \geq 1+\beta \log p$ for some constant $\beta > \frac{n-1}{\log 1/c}$ (lower bound in condition \ref{cond: degree cond}) and in addition $A$ is generic. Then, $\krank(A) \geq cp$, with probability at least $1-\gamma_2 p^{-\beta'}$ for constants $\beta' > 0$ and $\gamma_2 > 0$, specified in \eqref{eq:probability rate constants_beta'} and \eqref{eq:probability rate constants_gamma2}.
\end{theorem}

\subsection*{Acknowledgements}
The authors acknowledge useful discussions with Sina Jafarpour, Adel Javanmard, Alex Dimakis, Moses Charikar, Sanjeev Arora, Ankur Moitra and Kamalika Chaudhuri. A. Anandkumar is supported in part by  Microsoft Faculty Fellowship, NSF Career award CCF-1254106, NSF Award CCF-1219234, ARO Award W911NF-12-1-0404, and ARO YIP Award W911NF-13-1-0084. M. Janzamin is supported by  NSF Award CCF-1219234, ARO Award W911NF-12-1-0404  and ARO YIP Award W911NF-13-1-0084.

\renewcommand{\appendixpagename}{Appendix}
\appendixpage
\appendix

\section{Proof of Deterministic Identifiability Result (Theorem \ref{Thm:Identifiability based on A})} \label{Appendix:ProofThm_Identifiability based on A}
First, we show the identifiability result under an alternative set of conditions on the $n$-gram matrix, $A^\ngram$, and then, we show that the conditions of Theorem \ref{Thm:Identifiability based on A} are sufficient for these conditions to hold.

\subsection{Deterministic analysis based on $A^\ngram$} \label{sec:ident. conditions based on A^ngram}
In this section, the deterministic identifiability result based on conditions on the $n$-gram matrix, $A^\ngram$, is provided.

In the $n$-gram matrix, $A^\ngram \in \Rbb^{p^n \times q}$, redundant rows exist. If some row of $A^\ngram$ is indexed by $n$-tuple $(i_1,\dotsc,i_n) \in [p]^n$, then another row indexed by any permutation of the tuple $(i_1,\dotsc,i_n)$ has the same entries. 
Therefore, the number of distinct rows of $A^\ngram$ is at most ${p+n-1 \choose n}$.
In the following definition, we define a non-redundant version of $n$-gram matrix which is restricted to the (potentially) distinct rows.

\begin{definition}[Restricted $n$-gram matrix] \label{def:ngram matrix restricted}
For any matrix $A \in \R^{p \times q}$, restricted $n$-gram matrix $A_\Rest^\ngram \in \R^{s
\times q}$, $s = {p+n-1 \choose n}$, is defined as the restricted version of $n$-gram matrix $A^\ngram \in \R^{p^n \times q}$, where the redundant rows of $A^\ngram$ are removed, as explained above.
\end{definition}


\begin{condition}[Rank condition] \label{cond:rank}
The $n$-gram matrix $A^\ngram$ is full column rank.
\end{condition}


\begin{condition}[Graph expansion] \label{cond:expansion}
Let $G (V_h,V_o^{(n)};A^\ngram)$ denote the bipartite graph with vertex sets $V_h$ corresponding to the hidden variables (indexing the columns of $A^\ngram$) and $V_o^{(n)}$ corresponding to the $n$-th order observed variables (indexing the rows of $A^\ngram$) and edge matrix $A^\ngram \in \Rbb^{|V_o^{(n)}| \times |V_h|}$. 
The bipartite graph $G (V_h,V_o^{(n)};A^\ngram)$ satisfies the following \textnormal{expansion property} on the restricted version specified by $A_\Rest^\ngram$,
\begin{align} \label{eq:expansion Appendix}
\Bigl| N_{A_\Rest^\ngram}(S) \Bigr| \geq |S| + d_{\max} \Bigl( A^\ngram \Bigr), \quad \forall S \subseteq V_h, \ |S| > \krank(A),
\end{align}
where $d_{\max} \Bigl( A^\ngram \Bigr)$ is the maximum node degree in set $V_h$.
\end{condition}
\begin{remark} \label{remark:conditions modifications}
The expansion condition for the bag-of-words admixture model is provided in \eqref{eq:expansion n=1}, introduced in \cite{AnandkumarEtal:LinearBayesianLatent}.
The proposed expansion condition in \eqref{eq:expansion Appendix} is inherited from \eqref{eq:expansion n=1}, with two major modifications. First, the condition is appropriately generalized for our model which involves a graph with edges specified by the $n$-gram matrix, $A^\ngram$, as stated in \eqref{eq:4th order moment of x in terms of h}. Second, the expansion property \eqref{eq:expansion n=1}, proposed in \cite{AnandkumarEtal:LinearBayesianLatent}, needs to be satisfied for all subsets $S$ with size $|S| \geq 2$, which is a stricter condition than the one proposed here in \eqref{eq:expansion Appendix}, since we can have $\krank(A) \gg 2$. 
\end{remark}

The deterministic identifiability result based on the conditions on $A^\ngram$, is stated in the following theorem for $n \geq 2$, while $n=1$ case is addressed in Remarks \ref{remark:n=1 ident. result} and \ref{remark:conditions modifications}. 
The identifiability result relies on access to the $(2n)$-th order moment of observed variables $x_l, l \in [2n]$, defined in equation \eqref{eq:4th order moment of x definition} as
\begin{align*}
M_{2n}(x) := \Ebb \left[ (x_1 \otimes x_2 \otimes \dotsm \otimes x_{n})(x_{n+1} \otimes x_{n+2} \otimes \dotsm \otimes x_{2n})^\top \right] \in \Rbb^{p^{n} \times p^{n}}.
\end{align*}


\begin{theorem}[Generic identifiability under deterministic conditions on $A^\ngram$] \label{Thm:Identifiability}
Let $M^{(n)}_{2n}(x)$ (defined in equation \eqref{eq:4th order moment of x definition}) be the $(2n)$-th order moment of the $n$-persistent topic model described in Section \ref{sec:model}.
If the model satisfies conditions \ref{cond:non-degeneracy}, \ref{cond:rank} and \ref{cond:expansion}, then, for any $n \geq 2$, all the columns of population structure $A$ are \textnormal{generically identifiable} from $M^{(n)}_{2n}(x)$.
\end{theorem}

\bprf
Define $B := A^\ngram \in \Rbb^{p^n \times q}$. Then, the moment characterized in equation \eqref{eq:4th order moment of x in terms of h} can be written as $M^{(n)}_{2n}(x) = B \Ebb \left[ hh^\top \right] B^\top$. 
Since both matrices $\Ebb \left[ hh^\top \right]$ and $B$ have full column rank (from  conditions \ref{cond:non-degeneracy} and \ref{cond:rank}), the rank of $B \Ebb \left[ hh^\top \right] B^\top$ is $q$ where $q=O(p^n)$, and furthermore $\Col(B \Ebb \left[ hh^\top \right] B^\top) = \Col(B)$. Let $\mathcal{U} := \{u_1,\dotsc,u_q \} \in \Rbb^{p^n}$ be any basis of $\Col(B \Ebb \left[ hh^\top \right] B^\top)$ satisfying the following two properties:
\begin{enumerate}
\item[1)] $u_i$'s have the smallest $\ell_0$ norms. 
\item[2)] $u_i$'s have $q$ smallest (tensor) ranks in the $n$-th order tensor form, i.e., $U_i := \ten(u_i), i \in [q]$, have $q$ smallest ranks.
\end{enumerate}
Let the columns of matrix $B$ be $b_i$ for $\ i \in [q]$. Since all the $b_i$'s (which belong to $\Col(B \Ebb \left[ hh^\top \right] B^\top)$) are rank-1 in the $n$-th order tensor form (since $\ten(b_i) = a_i^{\circ n}$) and the number of non-zero entries in each of $b_i$'s is at most $d_{\max}(B)=d_{\max}(A)^n$, we conclude that
\begin{align} \label{u_i's properties}
\max_i \rank(\ten(u_i)) = 1 \quad \text{and} \quad \max_i \lVert u_i \rVert_0 \leq d_{\max}(B).
\end{align}
The above bounds are concluded from the fact that $b_i \in \Col(B \Ebb \left[ hh^\top \right] B^\top), \ i \in [q],$ and therefore the $\ell_0$ norm and the rank properties of $b_i$'s are upper bounds for the corresponding properties of basis vectors $u_i$'s (according to the proposed conditions for $u_i$'s). \\
Now, exploiting these observations and also the genericity of $A$ and the expansion condition \ref{cond:expansion}, we show that the basis vectors $u_i$'s are scaled columns of $B$.
Since $u_i$ for  $i \in [q],$ is a vector in the column space of $B$, it can be represented as $u_i = B v_i$ for some vector $v_i \in \Rbb^q$.
Equivalently, for any $i \in [q]$, $u_i = \sum_{j=1}^{q} v_i(j) b_j$ where $b_j = a_j^{\otimes n}$ is the $j$-th column of matrix $B$ and $v_i(j)$ is a scalar which is the $j$-th entry of vector $v_i$. Then, the tensor form of $u_i$ can be written as
\begin{align}
\ten(u_i) = \sum_{j=1}^{q} v_i(j) \ten(b_j)
= \sum_{j=1}^{q} v_i(j) \ten(a_j^{\otimes n})
= \sum_{j=1}^{q} v_i(j) a_j^{\circ n}
= [[\Diagn(v_i); \overbrace{A,\dotsc,A}^{n \operatorname{times}}]], \label{eq:tensor form of u_i}
\end{align}
where the last equality is based on the notation defined in Definition \ref{def:CP decomp}. We define $\widetilde{v}_i := [v_i(j)]_{j:v_i(j) \neq 0}$ as the vector which contains only the non-zero entries of $v_i$, i.e., $\widetilde{v}_i$ is the restriction of vector $v_i$ to its support. Therefore, $\widetilde{v}_i \in \Rbb^r$, where $r := \Vert v_i \Vert_0$. Furthermore, the matrix $\widetilde{A}_i := \{a_j : v_i(j) \neq 0 \} \in \Rbb^{p \times r}$ is defined as the restriction of $A$ to its columns corresponding to the support of $v_i$. Let $(\widetilde{a}_i)_j$ denote the $j$-th column of $\widetilde{A}_i$. According to these definitions, equation \eqref{eq:tensor form of u_i}  reduces to
\begin{align} \label{eq:tensor form of u_i restricted}
\ten(u_i) = [[\Diagn(\widetilde{v}_i); \overbrace{\widetilde{A}_i,\dotsc,\widetilde{A}_i}^{n \operatorname{times}}]] = \sum_{j=1}^{r} \widetilde{v}_i(j) [(\widetilde{a}_i)_j]^{\circ n},
\end{align}
which is derived by removing columns of $A$ corresponding to the zero entries in $v_i$. \\
Next, we rule out that $\lVert v_i \rVert_0 \geq 2$ under two cases ($2 \leq \lVert v_i \rVert_0 \leq \krank(A)$ and $\krank(A) < \lVert v_i \rVert_0 \leq q$), to conclude that $u_i$'s vectors are scaled columns of $B$.
\paragraph{Case 1: $2 \leq \lVert v_i \rVert_0 \leq \krank(A)$.} Here, the number of columns of $\widetilde{A}_i \in \Rbb^{p \times \lVert v_i \rVert_0}$ is less than or equal to $\krank(A)$ and therefore it is full column rank.
Since, all the components of CP representation in equation \eqref{eq:tensor form of u_i restricted} are full column rank\,\footnote{Note that for $n \geq 3$, this full rank condition can be relaxed by Kruskal's condition for uniqueness of CP decomposition~\cite{Kruskal:77} and its generalization to higher order tensors~\cite{Sidiropoulos2000:CPuniqueness}. Precisely, instead of saying $\rank\bigl( \widetilde{A}_i \bigr)=\krank\bigl( \widetilde{A}_i \bigr) = r$, it is only required to have $\krank\bigl( \widetilde{A}_i \bigr) \geq (2r+n-1)/n$ to argue the result of case 1. This only improves the constants involved in the final result.}, for any\,\footnote{Note that for $n=1$, since the (tensor) rank of any vector is 1, this analysis does not work.} $n \geq 2$, we have $\rank(\ten(u_i)) = r = \Vert v_i \Vert_0 > 1$, which contradicts the fact that $\max_i \rank(\ten(u_i)) = 1$ in \eqref{u_i's properties}.
\paragraph{Case 2: $\krank(A) < \lVert v_i \rVert_0 \leq q$.} Here, we first restrict the $n$-gram matrix $B$ to distinct rows, denoted by $B_\Rest$, as defined in Definition \ref{def:ngram matrix restricted}. Let $u'_i =  B_\Rest v_i$. Since $u'_i$ is the restricted version of $u_i$, we have
\begin{align*}
\lVert u_i \rVert_0 & \geq \lVert u'_i \rVert_0 = \lVert B_\Rest v_i \rVert_0 \\
& > \bigl| N_{B_\Rest}(\Supp(v_i)) \bigr| - |\Supp(v_i)| \\
& \geq d_{\max}(B),
\end{align*}
where the second inequality is from Lemma \ref{lem:genericity}, and the third inequality follows from the graph expansion property (condition \ref{cond:expansion}). This result contradicts the fact that $\max_i \lVert u_i \rVert_0 \leq d_{\max}(B)$ in \eqref{u_i's properties}. \\
\\
From above contradictions, $\lVert v_i \rVert_0 = 1$ and hence, columns of $B := A^\ngram$ are the scaled versions of $u_i$'s.
\eprf

The following lemma is useful in the proof of Theorem \ref{Thm:Identifiability}. The result proposed in this lemma is similar to the parameter genericity condition in \cite{AnandkumarEtal:LinearBayesianLatent}, but generalized for the $n$-gram matrix, $A^\ngram$. The lemma is proved on lines of the proof of Remark 2.2 in \cite{AnandkumarEtal:LinearBayesianLatent}.
\begin{lem} \label{lem:genericity}
If $A \in \Rbb^{p \times q}$ is generic, then the $n$-gram matrix $A^\ngram \in \mathbb{R}^{p^n \times q}$ satisfies the following property with Lebesgue measure one.
For any vector $v \in \Rbb^q$ with $\| v \|_0 \geq 2$, we have
\begin{align*}
\left\| A_\Rest^\ngram v \right\|_0 > \left| N_{A_\Rest^\ngram}(\Supp(v)) \right| - |\Supp(v)|,
\end{align*}
where for a set $S \subseteq [q]$, $N_{A^\ngram}(S):= \{ i \in [p]^n : A^\ngram (i,j) \neq 0 \ for \ some \ j \in S \}$.
\end{lem}

Here, we prove the result for the case of $n=2$. The proof can be easily generalized to larger $n$.

Let $A := M + Z$ be generic, where $M$ is an arbitrary matrix, perturbed by random continuous perturbations $Z$. Consider the 2-gram matrix $B := A \odot A \in \Rbb^{p^2 \times q}$ . It is shown that the restricted version of $B$, denoted by $\tilB := B_\Rest \in \Rbb^{\frac{p(p+1)}{2} \times q}$, satisfies the above genericity condition.
We first establish some definitions.

\begin{definition}\label{def:fully dense}
We call a vector \emph{fully dense} if all of its entries are non-zero.
\end{definition}

\begin{definition}\label{def:NSP}
We say a matrix has the \emph{Null Space Property (NSP)} if its null space
does not contain any fully dense vector.
\end{definition}

\begin{claim} \label{claim:nullspace}
Fix any $S \subseteq [q]$ with $|S| \geq 2$, and  set $R := N_{M_\Rest^{(2\operatorname{-gram})}}(S)$.
Let $\tilCC$ be a $|S| \times |S|$ submatrix of $\tilB_{R,S}$.
Then $\Pr(\text{$\tilCC$ has the NSP}) = 1$.
\end{claim}

\bprfof{Claim \ref{claim:nullspace}}
First, note that $\tilB$ can be expanded as
\begin{align*}
\tilB := (A \odot A)_\Rest = (M \odot M)_\Rest + \underbrace{(M \odot Z + Z \odot M)_\Rest + (Z \odot Z)_\Rest}_{:= U}.
\end{align*}
Let $s = |S|$ and let $\tilCC = [\tilde{c}_1|\tilde{c}_2|\dotsb|\tilde{c}_s]^\top$, where $\tilde{c}_i^\top$ is the $i$-th row of $\tilCC$.
Also, let $C := [c_1|{c}_2|\dotsb|{c}_s]^\top$ and $W :=
[w_1|w_2|\dotsb|w_s]^\top$ be the corresponding $|S| \times |S|$ submatrices of $M_\Rest^{(2\operatorname{-gram})}$ and $U$, respectively.
For each $i \in [s]$, denote by $\mathcal{N}_i$ the null space of the
matrix $\tilCC_i = [\tilde{c}_1|\tilde{c}_2|\dotsb|\tilde{c}_i]^\top$.
Finally let $\mathcal{N}_0 = \R^s$.
Then, $\mathcal{N}_0 \supseteq \mathcal{N}_1\supseteq \dotsb \supseteq
\mathcal{N}_s$.
We need to show that, with probability one, $\mathcal{N}_s$ does not
contain any fully dense vector.

If one of $\mathcal{N}_i, i \in [s],$ does not contain any full dense vector, the result is proved.
Suppose that $\mathcal{N}_i$ contains some fully dense vector $v$.
Since $C$ is a submatrix of $M^{(2\operatorname{-gram})}_{R,S}$, every row
${c}_{i+1}^\top$ of ${C}$ contains at least one non-zero entry.
Therefore,
\begin{align*}
v^\top \tilde{c}_{i+1}
& = \sum_{j \in [s]} v(j) \tilde{c}_{i+1}(j)
\\
& = \sum_{j \in [s] : {c}_{i+1}(j) \neq 0}
v(j) ({c}_{i+1}(j) + w_{i+1}(j)),
\end{align*}
where $\{ w_{i+1}(j) : j \in [s] \ \text{s.t.} \ {c}_{i+1}(j) \neq 0
\}$ are independent random variables, and moreover, they are independent of $\tilde{c}_1, \dotsc,\tilde{c}_i$ and thus of $v$.
By assumption on the distribution of the $w_{i+1}(j)$,
\begin{eqnarray}
\Pr\Biggl[
v \in \mathcal{N}_{i+1}
\bigg| \tilde{c}_1,\tilde{c}_2,\dotsc,\tilde{c}_i \Biggr]
=
\Pr\Biggl[
\sum_{j \in [s] : {c}_{i+1}(j) \neq 0}
v(j) ({c}_{i+1}(j) + w_{i+1}(j)) = 0
\bigg| \tilde{c}_1,\tilde{c}_2,\dotsc,\tilde{c}_i \Biggr]
= 0 .
\end{eqnarray}
Consequently,
\begin{eqnarray}
\Pr\Biggl[
\dim(\mathcal{N}_{i+1}) < \dim(\mathcal{N}_i) \bigg| \tilde{c}_1,\tilde{c}_2,\dotsc,\tilde{c}_i
\biggr] = 1
\end{eqnarray}
for all $i = 0,\dotsc, s-1$.
As a result, with probability one, $\dim(\mathcal{N}_s) = 0$.
\eprfof

Now, we are ready to prove Lemma \ref{lem:genericity}.

\bprfof{Lemma \ref{lem:genericity}}
It follows from Claim~\ref{claim:nullspace} that, with probability one, the
following event holds: for every $S \subseteq [q], |S| \geq 2$, and
every $|S| \times |S|$ submatrix $\tilCC$ of $\tilB_{R,S}$ where  $R := N_{M_\Rest^{(2\operatorname{-gram})}}(S)$, then $\tilde{C}$ has
the NSP.

Now fix $v \in \R^q$ with $\|v\|_0 \geq 2$.
Let $S := \Supp(v)$ and $H := \tilB_{R,S}$.
Furthermore, let $u \in (\R \setminus \{0\})^{|S|}$ be the restriction
of vector $v$ to $S$; observe that $u$ is fully dense.
It is clear that $\|\tilB v\|_0 = \|H u\|_0$, so we need to show that
\begin{eqnarray}
\|Hu\|_0 > |R| - |S| .
\end{eqnarray}
For the sake of contradiction, suppose that $Hu$ has at most $|R| - |S|$
non-zero entries. Since $Hu \in \Rbb^{|R|}$, there is a subset of $|S|$ entries on which $Hu$ is zero.
This corresponds to a $|S| \times |S|$ submatrix of $H :=  \tilB_{R,S}$ which contains
$u$ in its null space. It means that this submatrix does not have
the NSP, which is a contradiction.
Therefore we conclude that $Hu$ must have more than $|R| - |S|$
non-zero entries, which finishes the proof.
\eprfof

\subsection{Proof of moment characterization lemmata} \label{Appendix:proofoflem_persistent topic model moment characterization}
\begin{remark} \label{remark:moment lemma more derivaition}
In Lemmata \ref{lem:persistent topic model moment characterization} and \ref{lem:persistent topic model moment characterization tensor form},  a specific case of order and persistence ($m=rn$) was considered. Here, we provide the moment form for a more general case.
Assume that $m=rn+s$ for some integers $r \geq 1,1 \leq s \leq \frac{n}{2}$, then
\begin{align*}
M^{(n)}_{2m}(x) = & \biggl( \overbrace{A^\ngram  \otimes \dots \otimes A^\ngram}^{r \operatorname{\ times}}  \otimes A^{(s \operatorname{-gram})} \biggr) \\
& \widetilde{M}_{2r}(h) \biggl( A^{((n-s) \operatorname{-gram})} \otimes \overbrace{A^\ngram  \otimes \dots \otimes A^\ngram}^{r-1 \operatorname{\ times}} \otimes A^{(2s \operatorname{-gram})} \biggr)^\top,
\end{align*}
where $\widetilde{M}_{2r}(h) \in \Rbb^{q^{r+1} \times q^{r+1}}$ is the hidden moment as
\begin{align*}
\widetilde{M}_{2r}(h)_{\bigl((i_1,\dotsc, i_{r+1}),(j_1,\dotsc, j_{r+1})\bigr)} :=
\left\{\begin{array}{ll}
\Ebb[h_{i_1} \dotsm h_{i_r} h_{i_{r+1}}^2 h_{j_2} \dotsm h_{j_{r+1}}] & \operatorname{if} i_{r+1}=j_1, \\
0 & \operatorname{o.w}.
\end{array}\right.
\end{align*}
The tensor form is also characterized as
\begin{align*}
T^{(n)}_{2m}(x) = \Bigl[ \Bigl[ \widetilde{S}_r; \overbrace{A,A,\dotsc,A}^{2m \ \operatorname{times}} \Bigr] \Bigr] \nn,
\end{align*}
where $\widetilde{S}_r \in \bigotimes^{2m} \Rbb^q$ is the core tensor in  the above Tucker representation with the sparsity pattern as follows. Let $\bfi:=  (i_1,i_2,\dotsc,i_{2m})$. If
\begin{align*}
& i_1=i_2=\dotsb=i_n,
i_{n+1}=i_{n+2}=\dotsb=i_{2n},
\dotsb,
i_{(2r-1)n+1}=i_{(2r-1)n+2}=\dotsb=i_{2rn}, \\
& i_{2(m-s)+1}=i_{2(m-s)+2}=\dotsb=i_{2m},
\end{align*}
we have
\[
\widetilde{S}_r \bigl(\bfi \bigr) = \widetilde{M}_{2r}(h)_{\bigl((i_n, i_{2n},\dotsc, i_{rn}, i_m),(i_{(r+1)n},i_{(r+2)n},\dotsc, i_{2rn},i_{2m})\bigr)}. \]
Otherwise, $\widetilde{S}_r \bigl(\bfi \bigr) = 0$.
\end{remark}

\bprfof{Lemma \ref{lem:persistent topic model moment characterization}}
In order to simplify the notation, similar to tensor powers for vectors, the tensor power for a matrix $U \in \Rbb^{p \times q}$ is defined as
\begin{align} \label{eq:tensor power for matrix}
U^{\otimes r} := \overbrace{U  \otimes U \otimes \dots \otimes U}^{r \operatorname{\ times}} \in \Rbb^{p^r \times q^r}.
\end{align}
First, consider the case $m=rn$ for some integer $r \geq 1$. One advantage of encoding $y_j, j \in [2r]$, by basis vectors appears in characterizing the conditional moments. The first order conditional moment of words $x_l, l \in [2m]$, in the $n$-persistent topic model can be written as
\begin{align*}
\mathbb{E} \bigl[ x_{(j-1)n+k} | y_j \bigr] = A y_j, \ j \in [2r], \ k \in [n],
\end{align*}
where $A=[a_1|a_2|\dotsb|a_q] \in \Rbb^{p \times q}$.
Next, the $m$-th order conditional moment of different views $x_l, l \in [m]$, in the $n$-persistent topic model can be written as
\begin{align*}
\mathbb{E}[x_1 \otimes x_2 \otimes \dotsb \otimes x_m | y_1=e_{i_1},y_2=e_{i_2},\dotsc,y_r=e_{i_r}] = a_{i_1}^{\otimes n} \otimes a_{i_2}^{\otimes n} \otimes \dots \otimes a_{i_r}^{\otimes n},
\end{align*}
which is derived from the conditional independence relationships among the observations $x_l, l \in [m]$, given topics $y_j, j \in [r]$. 
Similar to the first order moments, since vectors $y_j, j \in [r]$, are encoded by the basis vectors $e_i \in \Rbb^q$, the above moment can be written as the following matrix multiplication
\begin{align} \label{eq:4th order conditional moment of x}
\mathbb{E}[x_1 \otimes x_2 \otimes \dotsb \otimes x_m | y_1,y_2,\dotsc,y_r] = \Bigl( A^\ngram  \Bigr)^{\otimes r} \left( y_1 \otimes y_2 \otimes \dots \otimes y_r \right),
\end{align}
where the $(\cdot)^{\otimes r}$ notation is defined in equation \eqref{eq:tensor power for matrix}.
Now for the $(2m)$-th order moment, we have
\begin{align}
M^{(n)}_{2m}(x) := \ & \Ebb \Big[ (x_1 \otimes x_2 \otimes \dotsm \otimes x_m)(x_{m+1} \otimes x_{m+2} \otimes \dotsm \otimes x_{2m})^\top \Bigr] \nn \\
= \ & \mathbb{E}_{(y_1,y_2,\dotsc,y_{2r})} \Bigl[ \mathbb{E} \left[ (x_1 \otimes \dotsm \otimes x_m)(x_{m+1} \otimes \dotsm \otimes x_{2m})^\top | y_1,y_2,\dotsc,y_{2r} \right] \Bigr] \nn \\
\stackrel{(a)}{=} \ & \mathbb{E}_{(y_1,y_2,\dotsc,y_{2r})} \Bigl[ \mathbb{E} \bigl[ (x_1 \otimes \dotsm \otimes x_m) | y_1,\dotsc,y_{2r} \bigr] \mathbb{E} \bigl[ (x_{m+1} \otimes \dotsm \otimes x_{2m})^\top | y_1,\dotsc,y_{2r} \bigr] \Bigr] \nn \\
\stackrel{(b)}{=} \ & \mathbb{E}_{(y_1,y_2,\dotsc,y_{2r})} \Bigl[ \mathbb{E} \bigl[ (x_1 \otimes \dotsm \otimes x_m) | y_1,\dotsc,y_r \bigr] \mathbb{E} \bigl[ (x_{m+1} \otimes \dotsm \otimes x_{2m})^\top | y_{r+1},\dotsc,y_{2r} \bigr] \Bigr] \nn \\
\stackrel{(c)}{=} \ & \mathbb{E}_{(y_1,y_2,\dotsc,y_{2r})} \left[ \biggl( \Bigl[ A^\ngram  \Bigr]^{\otimes r} \biggr) \left( y_1 \otimes \dots \otimes y_r \right) \left( y_{r+1} \otimes \dots \otimes y_{2r} \right)^\top \biggl( \Bigl[ A^\ngram  \Bigr]^{\otimes r} \biggr)^\top \right] \nn \\
= \ & \biggl( \Bigl[ A^\ngram  \Bigr]^{\otimes r} \biggr) \mathbb{E} \left[ \left( y_1 \otimes \dots \otimes y_r \right) \left( y_{r+1} \otimes \dots \otimes y_{2r} \right)^\top \right]  \biggl( \Bigl[ A^\ngram  \Bigr]^{\otimes r} \biggr)^\top \nn \\
\stackrel{(d)}{=} \ & \biggl( \Bigl[ A^\ngram  \Bigr]^{\otimes r} \biggr) M_{2r}(y)  \biggl( \Bigl[ A^\ngram  \Bigr]^{\otimes r} \biggr)^\top, \label{eq:4th order moment of x}
\end{align}
where $(a)$ results from the independence of $(x_1,\dotsc,x_m)$ and $(x_{m+1},\dotsc,x_{2m})$ given $(y_1,y_2,\dotsc,y_{2r})$ and $(b)$ is concluded from the independence of $(x_1,\dotsc,x_m)$ and $(y_{r+1},\dotsc,y_{2r})$ given $(y_1,\dotsc,y_r)$ and the independence of $(x_{m+1},\dotsc,x_{2m})$ and $(y_1,\dotsc,y_r)$ given $(y_{r+1},\dotsc,y_{2r})$.  Equation \eqref{eq:4th order conditional moment of x} is used in $(c)$ and finally, the $(2r)$-th order moment of $(y_1,\dotsc,y_{2r})$ is defined as $M_{2r}(y) := \mathbb{E} \left[ \left( y_1 \otimes \dots \otimes y_r \right) \left( y_{r+1} \otimes \dots \otimes y_{2r} \right)^\top \right]$ in $(d)$.

For $M_{2r}(y)$, we have by the law of total expectation
\begin{align*}
M_{2r}(y) := \ & \Ebb \bigl[ \left( y_1 \otimes \dots \otimes y_r \right) \left( y_{r+1} \otimes \dots \otimes y_{2r} \right)^\top \bigr] \\
= \ & \Ebb_h  \Bigl[ \Ebb \bigl[ \left( y_1 \otimes \dots \otimes y_r \right) \left( y_{r+1} \otimes \dots \otimes y_{2r} \right)^\top | h \bigr] \Bigr] \\
= \ & \Ebb_h  \biggl[  \Bigl( \overbrace{h \otimes \dotsm \otimes h}^{r \operatorname{times}} \Bigr)  \Bigr( \overbrace{h \otimes \dotsm \otimes h}^{r \operatorname{times}} \Bigl) ^\top \biggr] \\
= \ & M_{2r}(h),
\end{align*}
where the third equality is concluded from the conditional independence of variables $y_j, j \in [2r]$, given $h$ and the model assumption that $\Ebb \bigl[ y_j |h \bigr] = h, j \in [2r]$.
Substituting this in equation \eqref{eq:4th order moment of x}, finishes the proof for the $n$-persistent topic model. Similarly, the moment of single topic model (infinite persistence) can be also derived.
\eprfof

\bprfof{Lemma \ref{lem:persistent topic model moment characterization tensor form}}
Defining $\Lambda := M_{2r}(h) \in \Rbb^{q^r \times q^r}$ and $B := \bigl[ A^\ngram \bigr]^{\otimes r} \in \Rbb^{p^{rn} \times q^r}$, the $(2rn)$-th order moment $M^{(n)}_{2rn}(x) \in \Rbb^{p^{rn} \times p^{rn}}$ of the $n$-persistent topic model proposed in equation \eqref{eq:2m-th order moment of x in terms of h} can be written as
\begin{align*}
M^{(n)}_{2rn}(x) = B \Lambda B^\top.
\end{align*}
Let $b_{(i_1,\dotsc,i_r)} \in \Rbb^{p^{rn}}$ denote the corresponding column of $B$ indexed by $r$-tuple $(i_1,\dotsc,i_r), i_k \in [q], k \in [r]$. 
Then, the above matrix equation can be expanded as
\begin{align*}
M^{(n)}_{2rn}(x)
& = \sum_{\substack{i_1,\dotsc,i_r \in [q] \\ j_1,\dotsc,j_r \in [q]}} \Lambda \bigl( (i_1,\dotsc,i_r),(j_1,\dotsc,j_r) \bigr) b_{(i_1,\dotsc,i_r)} b_{(j_1,\dotsc,j_r)}^\top \\
& = \sum_{\substack{i_1,\dotsc,i_r \in [q] \\ j_1,\dotsc,j_r \in [q]}} \Lambda \bigl( (i_1,\dotsc,i_r),(j_1,\dotsc,j_r) \bigr) [a_{i_1}^{\otimes n} \otimes \dots \otimes a_{i_r}^{\otimes n}] [a_{j_1}^{\otimes n} \otimes \dots \otimes a_{j_r}^{\otimes n}]^\top,
\end{align*}
where relation $b_{(i_1,\dotsc,i_r)} = a_{i_1}^{\otimes n} \otimes \dots \otimes a_{i_r}^{\otimes n}, i_1,\dotsc,i_r \in [q],$ is used in the last equality. Let $m^{(n)}_{2rn}(x) \in \Rbb^{p^{2rn}}$ denote the vectorized form of $(2rn)$-th order moment $M^{(n)}_{2rn}(x) \in \Rbb^{p^{rn} \times p^{rn}}$. Therefore, we have
\begin{align*}
m^{(n)}_{2rn}(x) := & \vecform \Bigl( M^{(n)}_{2rn}(x) \Bigr) \\
 = & \sum_{\substack{i_1,\dotsc,i_r \in [q] \\ j_1,\dotsc,j_r \in [q]}} \Lambda \bigl( (i_1,\dotsc,i_r),(j_1,\dotsc,j_r) \bigr) a_{i_1}^{\otimes n} \otimes \dots \otimes a_{i_r}^{\otimes n} \otimes a_{j_1}^{\otimes n} \otimes \dots \otimes a_{j_r}^{\otimes n}.
\end{align*}
Then, we have the following equivalent tensor form for the original model proposed in equation \eqref{eq:2m-th order moment of x in terms of h}
\begin{align*}
T^{(n)}_{2rn}(x) := & \ten \Bigl( m^{(n)}_{2rn}(x) \Bigr) \\
= & \sum_{\substack{i_1,\dotsc,i_r \in [q] \\ j_1,\dotsc,j_r \in [q]}} \Lambda \bigl( (i_1,\dotsc,i_r),(j_1,\dotsc,j_r) \bigr) a_{i_1}^{\circ n} \circ \dots \circ a_{i_r}^{\circ n} \circ a_{j_1}^{\circ n} \circ \dots \circ a_{j_r}^{\circ n}.
\end{align*}
\eprfof

\subsection{Sufficient matching properties for satisfying rank and graph expansion conditions} \label{Appendix:lem_suff matching for rank and expansion}
In the following lemma, it is shown that under a perfect $n$-gram matching and additional genericity and krank conditions, the rank and graph expansion conditions \ref{cond:rank} and \ref{cond:expansion} on $A^\ngram$, are satisfied.

\begin{lem} \label{lem:suff matching for rank and expansion}
Assume that the bipartite graph $G(V_h,V_o;A)$ has a perfect $n$-gram matching (condition \ref{cond:perfect ngram matching} is satisfied). Then, the following results hold for the $n$-gram matrix $A^\ngram$:
\begin{itemize}
\item[1)] If $A$ is generic, $A^\ngram$ is full column rank (condition \ref{cond:rank}) with Lebesgue measure one (almost surely).
\item[2)] If krank condition \ref{cond:krank bound} holds, $A^\ngram$ satisfies the proposed expansion property in condition \ref{cond:expansion}. 
\end{itemize}
\end{lem}

\bprf
Let $M$ denote the perfect $n$-gram matching of the bipartite graph $G(V_h,V_o;A)$. From Lemma \ref{lem:ngram perfect matching property}, there exists a perfect matching $M^\ngram$ for the bipartite graph $G(V_h,V_o^{(n)};A^{\ngram})$.  Denote the corresponding bi-adjacency matrix to the edge set $M$ as $A_M$. Similarly, $B_M$ denotes the corresponding bi-adjacency matrix to the edge set $M^\ngram$. 
Note that $\Supp(A_M) \subseteq \Supp(A)$ and $\Supp(B_M) \subseteq \Supp(A^\ngram)$.

Since $B_M$ is a perfect matching, it consists of $q := |V_h|$ rows, each of which has only one non-zero entry, and furthermore, the non-zero entries are in $q$ different columns. Therefore, these rows form $q$ linearly independent vectors. Since the row rank and column rank of a matrix are equal, and the number of columns of $B_M$ is $q$, the column rank of $B_M$ is $q$ or in other words, $B_M$ is full column rank. Since $A$ is generic, from Lemma \ref{lem:full column rank submatrix} (with a slight modification in the analysis\,\footnote{Lemma \ref{lem:full column rank submatrix} result is about the column rank of $A$ itself, but here it is about the column rank of $A^\ngram$ for which the same analysis works. Note that the support of $B_M$ (which is full column rank here) is within the support of $A^\ngram$ and therefore Lemma \ref{lem:full column rank submatrix} can still be applied.}), $A^\ngram$ is also full column rank with Lebesgue measure one (almost surely). This completes the proof of part 1.

Next, the second part is proved.  From krank definition, we have
\begin{align*}
|N_A(S')| \geq |S'|  \quad  \operatorname{for} \ S' \subseteq V_h, |S'| \leq \krank(A),
\end{align*}
which is concluded from the fact that the corresponding submatrix of $A$ specified by $S'$ should be full column rank.
From this inequality, we have
\begin{align} \label{eq:proof aux}
|N_A(S')| \geq \krank(A)  \quad  \operatorname{for} \ S' \subseteq V_h, |S'| = \krank(A).
\end{align}
Then, we have
\begin{align}
|N_A(S)| & \geq |N_A(S')| \quad  \operatorname{for} \ S' \subset S \subseteq V_h, |S| > \krank(A), |S'| = \krank(A), \nn \\
& \geq \krank(A) \nn \\
& \geq d_{\max}(A)^n, \label{eq:aux expansion}
\end{align}
where \eqref{eq:proof aux} is used in the second inequality and the last inequality is from krank condition \ref{cond:krank bound}.

In the restricted $n$-gram matrix $A_\Rest^\ngram$, the number of neighbors for a set $S \subseteq V_h, |S| > \krank(A)$, can be bounded as
\begin{align*}
\left| N_{A_\Rest^\ngram}(S) \right| & \geq |N_A(S)| + |S|
 \\
& \geq  d_{\max} (A)^n+|S|   \quad  \operatorname{for} \ |S|>\krank(A),
\end{align*}
where the first inequality is due to the fact that the set $N_{A_\Rest^\ngram}$ consists of rows indexed by the following two subsets: $n$-tuples $(i,i,\ldots,i)$ where all the indices are equal and $n$-tuples $(i_1, \ldots, i_n)$ with distinct indices, i.e., $i_1\neq i_2\ldots \neq i_n$. The former subset is exactly $N_A(S)$ while the size of the latter subset is at least $|S|$ due to the existence of a perfect $n$-gram matching in $A$. The bound \eqref{eq:aux expansion} is used in the second inequality.
Since $d_{\max} \bigl(A^\ngram \bigr) = d_{\max}(A)^n$, the proof of part 2 is also completed.

\eprf

\begin{remark}
The second result of above lemma is similar to the necessity argument of (Hall's) Theorem \ref{Thm:Hall's theorem} for the existence of perfect matching in a bipartite graph, but generalized to the case of perfect $n$-gram matching and with additional krank condition. 
\end{remark}

\subsection{(Auxiliary) lemma} \label{Appendix:AuxLemma}

\bprfof{Lemma \ref{lem:ngram perfect matching property}}
We show that if $G(Y,X;A)$ has a perfect $n$-gram matching, then $G(Y,X^{(n)};A^{\ngram})$ has a perfect matching. The reverse can be also immediately shown by reversing the discussion and exploiting the additional condition stated in the lemma. \\
Let $E^\ngram$ denote the edge set of the bipartite graph $G(Y,X^{(n)};A^{\ngram})$.
Assume $G(Y,X;A)$ has a perfect $n$-gram matching $M \subseteq E$. For any $j \in Y$, let $N_M(j)$ denote the set of neighbors of vertex $j$ according to edge set $M$. Since $M$ is a perfect $n$-gram matching, $|N_M(j)| = n$ for all $j \in Y$. 
It can be immediately concluded from Definition \ref{def:$n$-gram Matching} that sets $N_M(j)$ are all distinct, i.e., $N_M(j_1) \neq N_M(j_2)$ for any  $j_1,j_2 \in Y, j_1 \neq j_2$. For any $j \in Y$, let $N'_M(j)$ denote an arbitrary ordered $n$-tuple generated from the elements of set $N_M(j)$.
From the definition of $n$-gram matrix, we have $A^\ngram(N'_M(j),j) \neq 0$ for all $j \in Y$. Hence, $(j,N'_M(j)) \in E^\ngram$ for all $j \in Y$ which together with the fact that all $N'_M(j)$'s tuples are distinct, it results that $M^\ngram := \{ (j,N'_M(j))|j \in Y \} \subseteq E^\ngram$ is a perfect matching for $G(Y,X^{(n)};A^{\ngram})$.
\eprfof


\begin{lem} \label{lem:full column rank submatrix}
Consider matrix $C \in \Rbb^{m \times r}$ which is generic. Let $\widetilde{C} \in \Rbb^{m \times r}$ be such that $\Supp(\widetilde{C}) \subseteq \Supp(C)$ and the non-zero entries of $\widetilde{C}$  are the same as the corresponding non-zero entries of $C$. If $\widetilde{C}$ is full column rank, then $C$ is also full column rank, almost surely.
\end{lem}
\bprf
Since $\widetilde{C}$ is full column rank, there exists a $r \times r$ submatrix of $\widetilde{C}$, denoted by $\widetilde{C}_S$, with non-zero determinant, i.e., $\det(\widetilde{C}_S) \neq 0$. Let $C_S$ denote the corresponding submatrix of $C$ indexed by the same rows and columns as $\widetilde{C}_S$. \\
The determinant of $C_S$ is a polynomial in the entries of $C_S$. Since $\widetilde{C}_S$ can be derived from $C_S$ by keeping the corresponding non-zero entries, $\det(C_S)$ can be decomposed into two terms as
\begin{align*}
\det(C_S) = \det(\widetilde{C}_S) + f(C_S),
\end{align*}
where the first term corresponds to the monomials for which all the variables (entries of $C_S$) are also in $\widetilde{C}_S$ and the second term corresponds to the monomials for which at least one variable is not in $\widetilde{C}_S$. The first term is non-zero as stated earlier. Since $C$ is generic, the polynomial $f(C_S)$ is non-trivial and therefore its roots have Lebesgue measure zero. It implies that $\det(C_S) \neq 0$ with Lebesgue measure one (almost surely), and hence, it is full (column) rank. Thus, $C$ is also full column rank, almost surely.
\eprf

Finally, Theorem \ref{Thm:Identifiability based on A} is proved by combining the results of Theorem \ref{Thm:Identifiability} and Lemma \ref{lem:suff matching for rank and expansion}.

\bprfof{Theorem \ref{Thm:Identifiability based on A}}
Since conditions \ref{cond:perfect ngram matching} and \ref{cond:krank bound} hold and $A$ is generic, Lemma \ref{lem:suff matching for rank and expansion} can be applied which results that rank condition \ref{cond:rank} is satisfied almost surely and expansion condition \ref{cond:expansion} also holds. Therefore, all the required conditions for Theorem \ref{Thm:Identifiability} are satisfied almost surely and this completes the proof.
\eprfof

\section{Proof of Random Identifiability Result (Theorem \ref{Thm:Identifiability random})} \label{Appendix:ProofThm_Identifiability random}
We provide detailed proof of the steps stated in the proof sketch of random result in Section \ref{sec:random proof sketch}. 

\subsection{Proof of existence of perfect $n$-gram matching and Kruskal results} \label{Appendix:ProofThm_Auxiliary Random Theorems}

\begin{figure} \centering
\bp
\psfrag{X}[l]{$X$}
\psfrag{X1}[l]{\small $X'_1$}
\psfrag{X2}[l]{\small $X'_2$}
\psfrag{Xn}[l]{\small $X'_n$}
\psfrag{X1U}[l]{\small $X_1$}
\psfrag{X2U}[l]{\small $X_2$}
\psfrag{XnU}[l]{\small $X_n$}
\psfrag{Y}[l]{$Y$}
\psfrag{Y1}[l]{\small $Y_1$}
\psfrag{Y2}[l]{\small $Y_2$}
\psfrag{YJ}[l]{\small $Y_J$}
\includegraphics[width=4.0in]{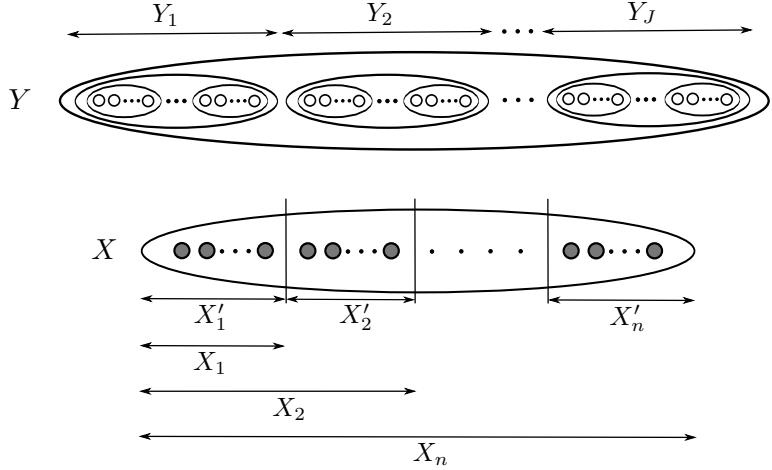}\ep
\caption{\small Partitioning of sets $Y$ and $X$, proposed in the proof of Theorem \ref{Thm:perfect n-gram matching random graph}. Set $X$ is randomly (uniform) partitioned  into $n$ sets of (almost) equal size, denoted by $X'_l, l \in [n]$. Set $Y$ is also randomly partitioned in a recursive manner. In each step, it is partitioned to $J = c \frac{p}{n} = O(p)$ number of sets. These smaller sets are again partitioned, recursively. This partitioning process is performed until reaching sets with size $O(p)$. The first two steps are shown in this figure.}
\label{fig:proof_Thm_partition}
\end{figure}

\bprfof{Theorem \ref{Thm:perfect n-gram matching random graph}}
Vertex sets $X$ and $Y$ are partitioned, described as follows (see Figure \ref{fig:proof_Thm_partition}).
Define $J := c \frac{p}{n}$. Partition set $X$ uniformly at random into $n$ sets of (almost) equal size\,\footnote{By almost, we mean the maximum difference in the size of partitions is 1 which is always possible.}, denoted by $X'_l, l \in [n]$. Define sets $X_l := \cup_{i=1}^l X'_i, l \in [n]$. Furthermore, partition set $Y$ uniformly at random, hierarchically as follows. First, partition into $J$ sets, each with size at most $\left( c \frac{p}{n} \right)^{n-1}$, and denote them by $Y_i, i \in [J]$. Next, partition each of these new smaller sets $Y_i$ further into $J$ sets, each with size at most $\left( c \frac{p}{n} \right)^{n-2}$. Do it iteratively up to $n-1$ steps, where at the end, set $Y$ is partitioned into sets with size at most $c \frac{p}{n}$. The first two steps are shown in Figure \ref{fig:proof_Thm_partition}.

\begin{figure} \centering
\begin{minipage}[b]{2.4in}
\bp
\psfrag{X}[l]{$Y$}
\psfrag{X1}[l]{\small $Y_1$}
\psfrag{X2}[l]{\small $Y_2$}
\psfrag{XJl}[l]{\small $Y_{J}$}
\psfrag{Y}[l]{$X$}
\psfrag{Y1}[l]{\small $X_{n-1}$}
\psfrag{Y2}[l]{\small $X'_n$}
\psfrag{G1}[l]{\small $M_1$}
\psfrag{G2}[l]{\small $M_2$}
\psfrag{GJl}[l]{\small $M_{J}$}
\centering \includegraphics[width=2.4in]{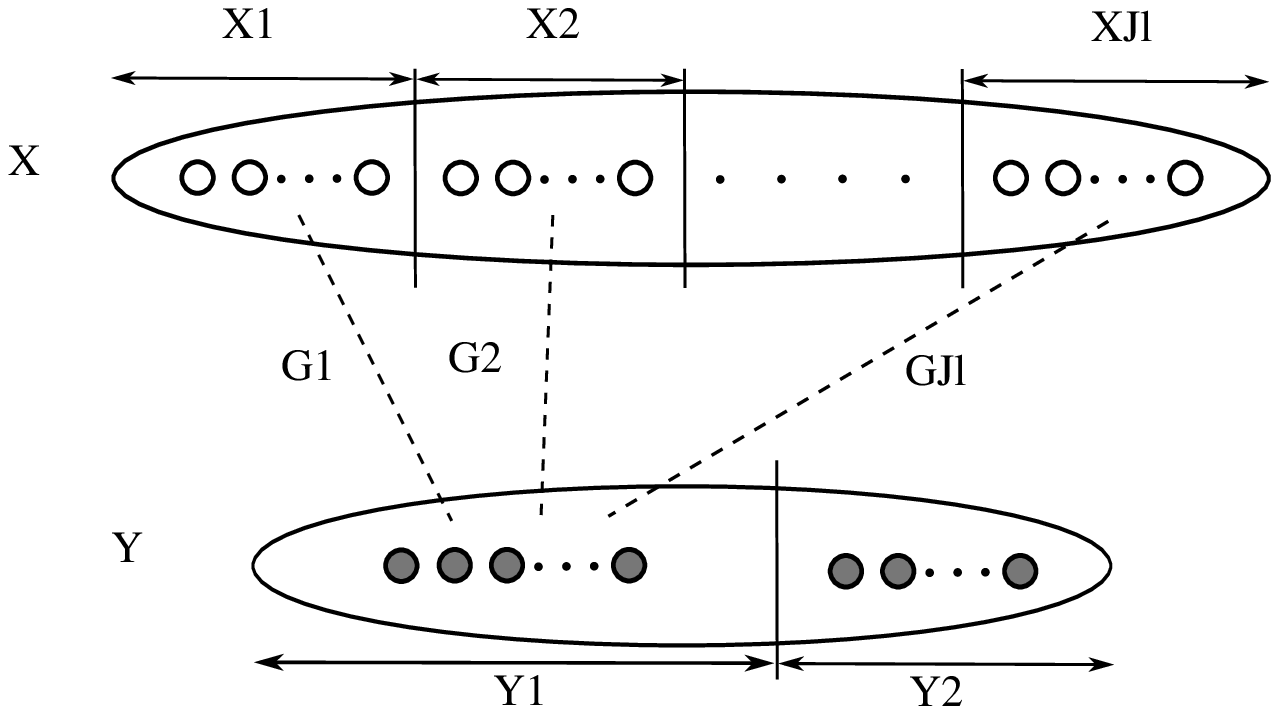}\ep
\subcaption{\small Partitioning of sets $Y$ and $X$ proposed for the induction step.} \label{fig:proof_lemma_partition1}
\end{minipage}
\hfil
\begin{minipage}[b]{2.4in}
\bp
\psfrag{X}[l]{$Y$}
\psfrag{Pa(S1)}[l]{\tiny $\operatorname{Pa}(S_1)$}
\psfrag{Pa(S2)}[l]{\tiny $\operatorname{Pa}(S_2)$}
\psfrag{Pa(S3)}[l]{\tiny $\operatorname{Pa}(S_3)$}
\psfrag{Y}[l]{$X$}
\psfrag{Y1}[l]{\small $X_{n-1}$}
\psfrag{Y2}[l]{\small $X'_n$}
\psfrag{S1}[l]{\tiny $S_1$}
\psfrag{S2}[l]{\tiny $S_2$}
\psfrag{S3}[l]{\tiny $S_3$}
\psfrag{PM}[l]{\tiny perfect matchings from $\operatorname{Pa}(S)$ to $X'_n$}
\centering \includegraphics[width=2in]{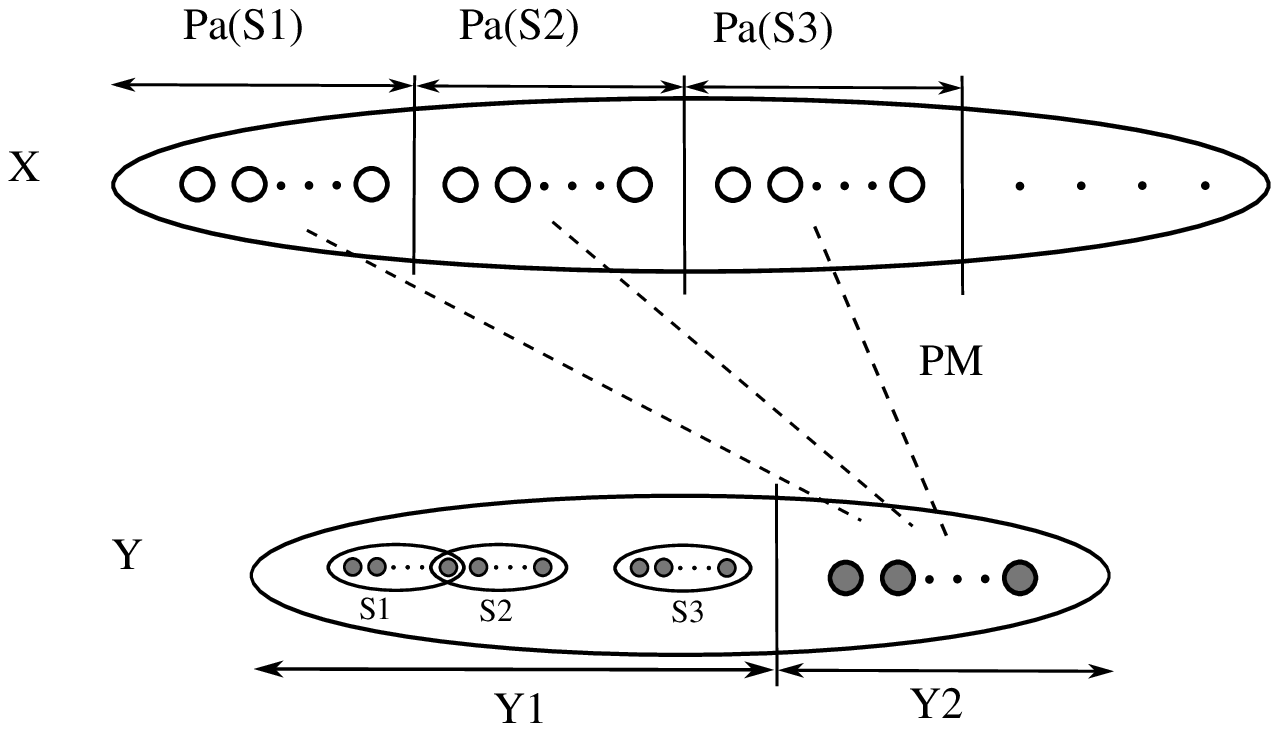}\ep
\subcaption{\small Partitioning of set $Y$ through perfect $(n-1)$-gram matchings $M_i, i \in [J]$.} \label{fig:proof_lemma_partition2}
\end{minipage}
\caption{\small Auxiliary figures for proof of induction step. (a) Partitioning of sets $Y$ and $X$ proposed in the proof, where set $Y$ is partitioned to $J := c \frac{p}{n}$ partitions $Y_1,\dotsc,Y_{J}$ with (almost) equal size, for some constant $c < 1$. In addition, set $X$ is partitioned to two partitions $X_{n-1}$ and $X'_n$ with sizes $|X_{n-1}| = \frac{n-1}{n} p$ and $|X'_n| = \frac{p}{n}$. The perfect $(n-1)$-gram matchings $M_i,  i \in [J],$ through bipartite graphs $G_i(Y_i,X_{n-1};E_i), i \in [J]$, are also highlighted in the figure. (b) Set $Y$ is partitioned to subsets $\operatorname{Pa}(S), S \in P_{n-1}(X_{n-1})$, which is generated through perfect $(n-1)$-gram matchings $M_i, i \in [J]$. $S_1$, $S_2$ and $S_3$ are three different sets in $P_{n-1}(X_{n-1})$ shown as samples. In addition, the perfect matchings from $\operatorname{Pa}(S), S \in P_{n-1}(X_{n-1})$, to $X'_n$, proposed in the proof, are also highlighted in the figure.}
\label{fig:proof_lemma}
\end{figure}

\paragraph{Proof by induction:}The existence of perfect $n$-gram matching from set $Y$ to set $X$ is proved by an induction argument. 
Consider one of intermediate sets in the hierarchical partitioning of $Y$ with size $O(p^l)$ and its further partitioning into $J := c \frac{p}{n}$ sets, each with size $O(p^{l-1})$, for any $l \in \{2,\dotsc,n\}$. In the induction step, it is shown that if there exists a perfect $(l-1)$-gram matching from each of these subsets of $Y$ with size $O(p^{l-1})$ to $X_{l-1}$, then there exists a perfect $l$-gram matching from the original set with size $O(p^l)$ to set $X_l$.
Specifically, in the last induction step, it is shown that if there exists a perfect $(n-1)$-gram matching from each set $Y_l, l \in [J],$ to set $X_{n-1}$, then there exists a perfect $n$-gram matching from $Y$ to $X_n=X$. 
\paragraph{Base case:}The base case of induction argument holds as follows. By applying Lemma \ref{lem:degree concentration bound} and Lemma \ref{lem:perfect matching random graph}, there exists a perfect matching from each partition in $Y$ with size at most $c \frac{p}{n} = O(p)$ to set $X_1$, \textbf{whp}.
\paragraph{Induction step:}Consider $J$ different bipartite graphs $G_i(Y_i,X_{n-1};E_i), i \in [J]$, by considering sets $Y_i$ and $X_{n-1}$ and the corresponding subset of edges $E_i \subset E$ incident to them. See Figure \ref{fig:proof_lemma_partition1}.  The induction step  is to show that if each of the corresponding $J$ bipartite graphs $G_i(Y_i,X_{n-1};E_i), i \in [J]$, has a perfect $(n-1)$-gram matching, then \textbf{whp}, the original bipartite graph $G(Y,X;E)$ has a perfect $n$-gram matching.

Let us denote the corresponding perfect $(n-1)$-gram matching of $G_i(Y_i,X_{n-1};E_i)$ by $M_i$. Furthermore, the set of all subsets of $X_{n-1}$ with cardinality $n-1$ are denoted by $P_{n-1}(X_{n-1})$, i.e., $P_{n-1}(X_{n-1})$ includes the sets with $(n-1)$ elements in the power set\,\footnote{The power set of any set $S$ is the set of all subsets of $S$.} of $X_{n-1}$.
For each set $S \in P_{n-1}(X_{n-1})$, take the set of all nodes in $Y$ which are connected to all members of $S$ according to the union of matchings $\cup_{i=1 }^{J} M_i$. Call this set as the parents of $S$, denoted by $\operatorname{Pa}(S)$.
According to the definition of perfect $(n-1)$-gram matching, there is at most one node in each set $Y_i$ which is connected to all members of $S$ through the matching $M_i$ and therefore, $|\operatorname{Pa}(S)| \leq J = c \frac{p}{n}$. In addition, note that sets $\operatorname{Pa}(S)$ impose a partitioning on set $Y$, i.e., each node $j \in Y$ is exactly included in one set $\operatorname{Pa}(S)$ for some $S \in P_{n-1}(X_{n-1})$. This is because of the perfect $(n-1)$-gram matchings considered for sets $Y_i, i \in [J]$. \\
Now, a perfect $n$-gram matching for the original bipartite graph is constructed as follows. For any $S \in P_{n-1}(X_{n-1})$, consider the set of parents $\operatorname{Pa}(S)$. Create the bipartite graph $G_S(\operatorname{Pa}(S),X'_n;E_S)$, where $E_S \subset E$ is the subset of edges incident to partitions $\operatorname{Pa}(S) \subset Y$ and $X'_n \subset X$. Denote by $d_S$ the minimum degree of nodes in set $\operatorname{Pa}(S)$ in the bipartite graph $G_S(\operatorname{Pa}(S),X'_n;E_S)$. Applying Lemma \ref{lem:degree concentration bound}, we have
\begin{align} \label{eq:degree concentration for matching}
\Pr[d_S \geq 1+ \beta \log (p/n)] & \geq 1 - J \exp \biggl( -\frac{2}{n^2} \frac{(d_{\min}-\beta n \log (p/n))^2}{d_{\min}} \biggr) \\
& \geq 1- \frac{c}{n} p^{-\beta \log 1/c} = 1-O(p^{-\beta \log 1/c}), \nn
\end{align}
where $\beta \log 1/c > n-1$, and the last inequality is concluded from the degree bound $d_{\min} \geq \alpha \log p$.
Furthermore, we have $|\operatorname{Pa}(S)| \leq c \frac{p}{n} = c |X'_n|$. Now, we can apply Lemma \ref{lem:perfect matching random graph} concluding that there exists a perfect matching from $\operatorname{Pa}(S)$ to $X'_n$ within the bipartite graph $G_S(\operatorname{Pa}(S),X'_n;E_S)$, with probability at least $1 - O(p^{-\beta \log 1/c})$. Refer to Figure \ref{fig:proof_lemma_partition2} for a schematic picture. The edges of this perfect matching are combined with the corresponding edges of the existing perfect $(n-1)$-gram matchings $M_i, i \in [J]$, to provide $n$ incident edges to each node $i \in \operatorname{Pa}(S)$. It is easy to see that this provides a perfect $n$-gram matching from $\operatorname{Pa}(S)$ to $X$. \\
We perform the same steps for all sets $S \in P_{n-1}(X_{n-1})$ to obtain a perfect $n$-gram matching from any $\operatorname{Pa}(S),S \in P_{n-1}(X_{n-1}),$ to $X$. Finally, according to this construction, the union of all of these matchings is a perfect $n$-gram matching from $\cup_{S \in P_{n-1}(X_{n-1})} \operatorname{Pa}(S) = Y$ to $X$. This finishes the proof of induction step. Note that here we analyzed the last induction step where the existence of perfect $n$-gram matching is concluded from the existence of corresponding perfect $(n-1)$-gram matchings. The earlier induction steps, where the existence of perfect $l$-gram matching is concluded from the existence of corresponding perfect $(l-1)$-gram matchings for any $l \in \{2,\dotsc,n\}$, can be similarly proven.
\paragraph{Probability rate:}We now provide the probability rate of the above events.
Let $N^{\text{(hp)}}_l, l \in [n]$, denote the total number of times that perfect matching result of Lemma \ref{lem:perfect matching random graph} is used in step $l$ in order to ensure that there exists a perfect $l$-gram matching from corresponding partitions of $Y$ to set $X_{l}$, \textbf{whp}. Let $N^{\text{(hp)}} = \sum_{l \in [n]} N^{\text{(hp)}}_l$. As earlier, let $P_{l-1} \bigl( X_{l-1} \bigr)$ denote the set of all subsets of $X_{l-1}$ with cardinality $l-1$. We have
\begin{align*}
\bigl| P_{l-1} \bigl( X_{l-1} \bigr) \bigr| = \binom{\bigl| X_{l-1} \bigr|}{l-1} = { \frac{l-1}{n}p \choose l-1}, \quad l \in \{ 2,\dotsc,n \}.
\end{align*}
According to the construction method of $l$-gram matching from $(l-1)$-gram matchings, proposed in the induction step, $\bigl| P_{l-1} \bigl( X_{l-1} \bigr) \bigr|$ is the number of times Lemma \ref{lem:perfect matching random graph} is used in order to ensure that there exists a perfect $l$-gram matching for each partition on the $Y$ side. Since at most $J^{n-l}$ number of such $l$-gram matchings are proposed in step $l$, the number $N^{\text{(hp)}}_l$ can be bounded as
\begin{align} \label{eq:N_l^hp bound}
N^{\text{(hp)}}_l \leq J^{n-l} \ \bigl| P_{l-1} \bigl( X_{l-1} \bigr) \bigr| =  J^{n-l} {\frac{l-1}{n}p \choose l-1}, \quad l \in \{ 2,\dotsc,n \}.
\end{align}
Since in the first step, $N^{\text{(hp)}}_1 = J^{n-1}$ number of perfect matchings needs to exist in the above discussion, we have
\begin{align*}
N^{\text{(hp)}} & = J^{n-1} + \sum_{l=2}^n N^{\text{(hp)}}_l \\
& \leq J^{n-1} + \sum_{l=2}^n  J^{n-l} {\frac{l-1}{n}p \choose l-1} \\
& \leq \Bigl( c \frac{p}{n} \Bigr)^{n-1} + \sum_{l=2}^n  \Bigl( c \frac{p}{n} \Bigr)^{n-l}  \Bigl( e \frac{p}{n} \Bigr)^{l-1} \\
& \leq n \Bigl( e \frac{p}{n} \Bigr)^{n-1} = O(p^{n-1}),
\end{align*}
where inequality \eqref{eq:N_l^hp bound} is used in the first inequality and $J := c \frac{p}{n}$ and inequality $\binom{n}{k} \leq \bigl( e \frac{n}{k} \bigr)^k$ are exploited in the second inequality. \\
Since the result of Lemma \ref{lem:perfect matching random graph} holds with probability at least $1 - O(p^{-\beta \log 1/c})$ and it is assumed that $\beta \log 1/c > n-1$, by applying union bound, we have the existence of perfect $n$-gram matching with probability at least $1-O(p^{-\beta'})$, for $\beta' = \beta  \log \frac{1}{c} - (n-1) > 0$. \\
Furthermore, note that the degree concentration bound in \eqref{eq:degree concentration for matching} is also used $O(p^{n-1})$ times. Since the bound in \eqref{eq:degree concentration for matching} holds with probability at least $1 - O(p^{-\beta \log 1/c})$ and it is assumed that $\beta \log 1/c > n-1$, this also reduces to the same probability rate. \\
The coefficient of the above polynomial probability rate is also explicitly computed, saying that the perfect $n$-gram matching exists with probability at least $1-\gamma_1 p^{-\beta'}$, with
\begin{align*}
\gamma_1 = e^{n-1} \Bigl( \frac{c}{n^{n-1}} + \frac{e^2}{1-\delta_1} n^{\beta'+1} \Bigr),
\end{align*}
where $\delta_1$ is a constant satisfying $e^2 \Bigl( \frac{p}{n} \Bigr)^{- \beta \log 1/c} < \delta_1 <1$.
\eprfof



\bprfof{Theorem \ref{Thm:krank bound for random case}}
Let $G(Y,X;A)$ denote the corresponding bipartite graph to matrix $A$ where node sets $Y = [q]$ and $X = [p]$ index the columns and rows of $A$ respectively. Therefore, $|Y| = q$ and $|X| = p$. Fix some $S \subseteq Y$ such that $|S| \leq p$.
Then
\begin{align} 
\Pr( |N(S)| \leq |S| )
& \leq \sum_{\substack{T \subseteq X : \\ |T|= |S|}} \Pr( N(S) \subseteq T ) \nn \\
& = \sum_{\substack{T \subseteq X : \\ |T|= |S|}} \prod_{i \in S} \binom{|S|}{d_i} \Bigl/ \binom{p}{d_i} \nn \\
& \leq \sum_{\substack{T \subseteq X : \\ |T|= |S|}} \prod_{i \in S} \biggl( \frac{|S|}{p} \biggr)^{d_i} \nn \\
& \leq \sum_{\substack{T \subseteq X : \\ |T|= |S|}} \prod_{i \in S} \biggl( \frac{|S|}{p} \biggr)^{d_{\min}} \nn \\
& = \binom{p}{|S|} \biggl( \frac{|S|}{p} \biggr)^{d_{\min}|S|}, \label{eq:neighbor bound prob}
\end{align}
where the bound $\binom{|S|}{d_i} \bigl/ \binom{p}{d_i} \leq  \Bigl( \frac{|S|}{p} \Bigr)^{d_i}$ is used in the second inequality, and the last inequality is concluded from the fact that $\frac{|S|}{p} \leq 1$.\\
Let $\mathcal{E}$ denote the event that for any subset $S \subseteq Y$ with $|S| \leq r$, we have $|N(S)| \geq |S|$, i.e.,
\begin{align*}
\mathcal{E} := ``\forall S \subseteq Y \wedge 1 \leq |S| \leq r : |N(S)| \geq |S|".
\end{align*}
Then, by the union bound and inequality \eqref{eq:neighbor bound prob}, we have
\begin{align*}
\Pr(\mathcal{E}^c) = \Pr( \exists S \subseteq Y \operatorname{s.t.} 1 \leq |S| \leq r \wedge |N(S)| < |S|)
& \leq \sum_{s=1}^r \binom{q}{s} \binom{p}{s} \biggl( \frac{s}{p} \biggr)^{d_{\min}s} \\
& \leq \sum_{s=1}^r
\biggl( e \frac{q}{s} \biggr)^s
\biggl( e \frac{p}{s} \biggr)^s
\biggl( \frac{s}{p} \biggr)^{d_{\min}s} \\
& \leq \sum_{s=1}^r \biggl( \frac{e^2 q r^{d_{\min}-2}}{p^{d_{\min}-1}} \biggr)^s,
\end{align*}
where the bound $\binom{n}{k} \leq \bigl( e \frac{n}{k} \bigr)^k$ is used in the second inequality. For $r = cp$ , the above inequality reduces to
\begin{align*}
\Pr(\mathcal{E}^c)
& \leq \sum_{s=1}^r \biggl( e^2 c^{d_{\min}-2} \frac{q}{p} \biggr)^s \\
& \leq \sum_{s=1}^r \Bigl( e^2 c' c^{d_{\min}-1} p^{n-1} \Bigr)^s \\
& \leq \sum_{s=1}^r \Bigl( e^2 c' c^{\beta \log p} p^{n-1} \Bigr)^s \\
& = \sum_{s=1}^r \Bigl( e^2 c' p^{n-1 - \beta \log 1/c} \Bigr)^s \\
& \leq \frac{e^2 c'}{p^{\beta'} - e^2 c'} = O(p^{-\beta'}), \quad \text{for} \ \beta' = \beta  \log \frac{1}{c} - (n-1) > 0,
\end{align*}
where the size condition assumed in the theorem is used in the second inequality with $c' := \frac{1}{c} \bigl( \frac{c}{n} \bigr)^n$, and the degree condition is exploited in the third inequality. The last inequality is concluded from the geometric series sum formula for large enough $p$. \\
Then, Lemma \ref{lem:krank expansion relation} can be applied concluding that $\krank(A) \geq r = c p$, with probability at least $1-\gamma_2 p^{-\beta'}$ for  constants $\beta' = \beta  \log \frac{1}{c} - (n-1) > 0$ and $\gamma_2 > 0$ as
\begin{align*}
\gamma_2 = \frac{c^{n-1} e^2}{n^n(1 - \delta_2)},
\end{align*}
where $\delta_2$ is a constant satisfying $c' e^2 p^{-\beta'} < \delta_2 <1$.
\eprfof

\bprfof{Remark \ref{remark:Insuff. of union bound}}
Consider a random bipartite graph $G(Y,X;E)$ where for each node $i \in X$:
\begin{enumerate}
\item Neighbors $N(i) \subseteq X$ are picked uniformly at random among all size $d$ subsets of $X$.
\item Matching $M(i) \subseteq N(i)$ is picked uniformly at random among all size $n$ subsets of $N(i)$.
\end{enumerate}
Note that as long as $n \leq d$, the distribution of $M(i)$ is uniform over all size $n$ subsets of $X$. \\
Fix some pair $i, i' \in Y$.
Then
\[ \Pr( M(i) = M(i') ) = \binom{|X|}{n}^{-1} . \]
By the union bound,
\[ \Pr\Bigl( \exists i,i' \in Y, i \neq i' \operatorname{s.t.} M(i) = M(i')
\Bigr) \leq \binom{|Y|}{2} \binom{|X|}{n}^{-1} , \]
which is $\Theta(|Y|^2/|X|^n)$ when $n$ is constant.
Therefore, if $d \geq n$ and the size constraint $|Y| = O(|X|^s)$ for some $s < \frac{n}{2}$ is satisfied, then \textbf{whp}, there is no pair of nodes in set $Y$ with the same random $n$-gram matching. This concludes that the random bipartite graph has a perfect $n$-gram matching \textbf{whp}, under these size and degree conditions. \\
\eprfof

\subsection{(Auxiliary) lemmata}

\begin{lem}[Existence of perfect matching for random bipartite graphs] \label{lem:perfect matching random graph}
Consider a random bipartite graph $G(W,Z;E)$ with $|W| = w$ nodes on the left side and $|Z| = z$ on the right side, and each node $i \in W$ is randomly connected to $d_i$ different nodes in set $Z$. Let $d_w := \min_{i \in W} d_i$. Assume that it satisfies the size condition $w \leq c z$ for some constant $0<c<1$ and the degree condition $d_w \geq 1 + \beta \log z$ for some constant $\beta > 0$. Then, there exists a perfect matching in the random bipartite graph $G(W,Z;E)$ with probability at least $1 - O(z^{-\beta \log 1/c})$ where $ \beta \log \frac{1}{c} > 0$.
\end{lem}
\bprf
From Hall's theorem (Theorem \ref{Thm:Hall's theorem}), the existence of perfect matching for a bipartite graph is equivalent to occurrence of the following event
\begin{align*}
\widetilde{\mathcal{E}} := ``\forall S \subseteq W : |N(S)| \geq |S|".
\end{align*}
Similar to the analysis in the proof of Theorem \ref{Thm:krank bound for random case}, it is concluded from union bound
\begin{align*}
\Pr \bigl( \widetilde{\mathcal{E}}^c \bigr) = \Pr( \exists S \subseteq W \operatorname{s.t.} |N(S)| < |S|)
& \leq \sum_{s=1}^{w} \binom{w}{s} \binom{z}{s} \biggl( \frac{s}{z} \biggr)^{d_w s} \\
& \leq \sum_{s=1}^{w}
\biggl( e \frac{w}{s} \biggr)^s
\biggl( e \frac{z}{s} \biggr)^s
\biggl( \frac{s}{z} \biggr)^{d_w s} \\
& \leq \sum_{s=1}^{w} \biggl( \frac{e^2 {w}^{d_w-1}}{z^{d_w-1}} \biggr)^s \\
& \leq \sum_{s=1}^{w} \Bigl(e^2 c^{d_w-1} \Bigr)^s,
\end{align*}
where the bound $\binom{n}{k} \leq \bigl( e \frac{n}{k} \bigr)^k$ is used in the second inequality. From the assumed lower bound on the degree $d_w$ and the fact that $0<c<1$, we have
\begin{align*}
\Pr \bigl( \widetilde{\mathcal{E}}^c \bigr)
& \leq \sum_{s=1}^{w} \Bigl( e^2 c^{\beta \log z} \Bigr)^s
= \sum_{s=1}^{w} \Bigl( e^2 z^{\beta \log c} \Bigr)^s
 \leq \frac{e^2}{ z^{\beta \log \frac{1}{c}} - e^2}
\leq \frac{e^2}{1- \delta_1} z^{- \beta \log 1/c},
\end{align*}
where the second inequality is concluded from the geometric series sum formula for large enough $z$, and $\delta_1$ is a constant satisfying $e^2 z^{- \beta \log 1/c} < \delta_1 <1$.
\eprf

\begin{lem}[Degree concentration bound] \label{lem:degree concentration bound}
Consider a random bipartite graph $G(Y,X;E)$ with $|Y|=q$ and $|X|=p$, where each node $i \in Y$ is randomly connected to $d_i$ different nodes in set $X$. Let $Y' \subset Y$ be any subset\,\footnote{Note that $Y'$ need not to be uniformly chosen and the result is valid for any subset of nodes $Y' \subset Y$.} of nodes in $Y$ with size $|Y'| = q'$ and $X' \subset X$ be a random (uniformly chosen) subset of nodes in $X$ with size $|X'| = p'$. Create the new bipartite graph $G(Y',X';E')$ where edge set $E' \subset E$ is the subset of edges in $E$ incident to $Y'$ and $X'$.  Denote the degree of each node $i \in Y'$ within this new bipartite graph by $d'_i$. Let $d_{\min} := \min_{i \in Y} d_i$ and $d'_{\min} := \min_{i \in Y'} d'_i$.
Then, if $d_{\min} > r \frac{p}{p'}$ for a non-negative integer $r$, we have
\[
\Pr[d'_{\min} \geq r+1] \geq 1 - q' \exp \biggl( -2 (p'/p)^2 \frac{(d_{\min}-(p/p')r)^2}{d_{\min}} \biggr).
\]
\end{lem}

\bprf
For any $i \in Y'$, we have
\begin{equation}
\Pr[d'_i \leq r] = \sum_{j=0}^r {p' \choose j } { {p-p'} \choose {d_i-j}} \Bigl/ {p \choose d_i}, \nn
\end{equation}
where the inner term of summation is a hypergeometric distribution with parameters $p$ (population size), $p'$ (number of success states in the population), $d_i$ (number of draws) and $j$ is the hypergeometric random variable denoting number of successes. The following tail bound for the hypergeometric distribution is provided \cite{Chvatal_HypergemetricTail1979,Skala_HypergemetricTail2011}
\begin{equation} \label{eq:hypergeo tail bound}
\Pr[d'_i \leq r] \leq \exp (-2 t_i^2 d_i), \nn
\end{equation}
for $t_i>0$ given by $r = \bigl( \frac{p'}{p} - t_i  \bigr) d_i$. Note that assumption $d_{\min} > \frac{p}{p'} r$ in the lemma is equivalent to having $t_i > 0, i \in Y$. Considering the minimum degree, for any $i \in Y'$, we have
\begin{align*}
\Pr[d'_i \leq r] \leq \exp (-2 t^2 d_{\min}),
\end{align*}
for $t>0$ given by $r = \bigl( \frac{p'}{p} - t  \bigr) d_{\min}$. Substituting $t$ from this equation gives the following bound
\begin{equation} \label{eq:degree concentration node}
\Pr[d'_i \leq r] \leq \exp \biggl( -2 (p'/p)^2 \frac{(d_{\min}-(p/p')r)^2}{d_{\min}} \biggr).
\end{equation}
Finally, applying the union bound, we can prove the result as follows
\begin{align*}
\Pr[d'_{\min} \geq r+1] = & \Pr[\cap_{i=1}^{q'} \{d'_i \geq r+1 \}]  \nn \\
\geq & 1 - \sum_{i=1}^{q'} \Pr[d'_i \leq r] \nn \\
\geq & 1 - \sum_{i=1}^{q'} \exp \biggl( -2 (p'/p)^2 \frac{(d_{\min}-(p/p')r)^2}{d_{\min}} \biggr) \nn \\
= & 1 - q' \exp \biggl( -2 (p'/p)^2 \frac{(d_{\min}-(p/p')r)^2}{d_{\min}} \biggr),
\end{align*}
where the union bound is applied in the first inequality and the second inequality is concluded from \eqref{eq:degree concentration node}.
\eprf


A lower bound on the Kruskal rank of matrix $A$ based on a sufficient relaxed expansion property on $A$ is provided in the following lemma.

\begin{lem} \label{lem:krank expansion relation}
If $A$ is generic and the bipartite graph $G(Y,X;A)$ satisfies the relaxed\,\footnote{There is no $d_{\max}$ term in contrast to the expansion property proposed in condition \ref{cond:expansion}.} expansion property $|N(S)| \geq |S|$ for any subset $S \subseteq Y$ with $|S| \leq r$, then $\krank(A) \geq r$, almost surely.
\end{lem}

Before proposing the proof, we state the marriage or Hall's theorem which gives an equivalent condition for having a perfect matching in a bipartite graph.
\begin{theorem}[Hall's theorem, \cite{Hall:marriage1935}] \label{Thm:Hall's theorem}
A bipartite graph $G(Y,X;E)$ has $Y$-saturating matching if and only if for every subset $S \subseteq Y$, the size of the neighbors of $S$ is at least as large as $S$, i.e., $|N(S)| \geq |S|$.
\end{theorem}

\bprfof{Lemma \ref{lem:krank expansion relation}}
Denote the submatrix $A_{N(S),S}$ by $\widetilde{A}_S$, i.e., $\widetilde{A}_S := A_{N(S),S}$. Exploiting marriage or Hall's theorem, it is concluded that the bipartite graph $G(S,N(S);\widetilde{A}_S)$ has a perfect matching $M_S$ for any subset $S \subseteq Y$ such that $|S| \leq r$. Denote by $\widetilde{A}_{M_S}$ the corresponding matrix to this perfect matching edge set $M_S$, i.e., $\widetilde{A}_{M_S}$ keeps the non-zero entries of $\widetilde{A}_S$ on edge set $M_S$ and everywhere else, it is zero. Note that the support of $\widetilde{A}_{M_S}$ is within the support of $\widetilde{A}_S$. According to the definition of perfect matching, the matrix $\widetilde{A}_{M_S}$ is full column rank. From Lemma \ref{lem:full column rank submatrix}, it is concluded that $\widetilde{A}_S$ is also full column rank almost surely. This is true for any $\widetilde{A}_S$ with $S \subseteq Y$ and $|S| \leq r$, which directly results that $\krank(A) \geq r$, almost surely.
\eprfof

Finally, Theorem \ref{Thm:Identifiability random} is proved by exploiting the random results on the existence of perfect $n$-gram matching and Kruskal rank, provided in Theorems \ref{Thm:perfect n-gram matching random graph} and \ref{Thm:krank bound for random case}.

\bprfof{Theorem \ref{Thm:Identifiability random}}
We claim that if random conditions \ref{cond: size cond} and \ref{cond: degree cond} are satisfied, then deterministic conditions \ref{cond:perfect ngram matching} and \ref{cond:krank bound} hold \textbf{whp}. Then Theorem \ref{Thm:Identifiability based on A} can be applied and the proof is done. \\
From size and degree conditions, Theorem \ref{Thm:perfect n-gram matching random graph} can be applied, which implies that the perfect $n$-gram matching condition \ref{cond:perfect ngram matching} is satisfied with probability at least $1-\gamma_1 p^{-\beta'}$ for $\beta' = \beta  \log \frac{1}{c} - (n-1) > 0$.
The conditions required for Theorem \ref{Thm:krank bound for random case} also hold and by applying this theorem we have the bound $\krank(A) \geq cp$, with probability at least$1-\gamma_2 p^{-\beta'}$.
Combining this inequality with the upper bound on degree $d$ in condition \ref{cond: degree cond}, we conclude that krank condition \ref{cond:krank bound} is also satisfied \textbf{whp}.
Hence, all the conditions required for Theorem \ref{Thm:Identifiability based on A} are satisfied with probability at least $1-\gamma p^{-\beta'}$, where
\begin{align*}
\gamma = \gamma_1 + \gamma_2 = e^{n-1} \Bigl( \frac{c}{n^{n-1}} + \frac{e^2}{1-\delta_1} n^{\beta'+1} \Bigr) + \frac{c^{n-1} e^2}{n^n(1 - \delta_2)},
\end{align*}
and this completes the proof.
\eprfof

\section{Relationship to CP Decomposition Uniqueness Results} \label{Appendix:CP uniqueness}
In this section, we provide a more detailed comparison with some uniqueness results of overcomplete CP decomposition. Here, the following CP decomposition for the third order tensor $T \in \Rbb^{p \times s \times q}$ is considered,
\begin{align} \label{eq:CP decomp third order}
T = \sum_{i=1}^r a_i \circ b_i  \circ c_i,
\end{align}
where $A = [a_1 | \dots | a_r] \in \Rbb^{p \times r}, B = [b_1 | \dots | b_r] \in \Rbb^{s \times r}$ and $C = [c_1 | \dots | c_r] \in \Rbb^{q \times r}$. \\
The most important and general uniqueness result of CP, called Kruskal's condition, is provided in \cite{Kruskal:77}, where it is guaranteed that the above CP decomposition is unique if
\begin{align*}
\krank(A) +\krank(B) +\krank(C) \geq 2r+2.
\end{align*}
Since then, several works have analyzed the uniqueness of CP decomposition.
One set of works assume that one of the components, say $C$, is full column rank \cite{LathauwerSIAM2006,jiang2004kruskal}. it is shown in \cite{LathauwerSIAM2006}, for generic (fully dense) components $A,B$ and $C$, if $r \leq q$ and $r(r-1) \leq p(p-1)s(s-1)/2$, then the CP decomposition in \eqref{eq:CP decomp third order} is generically unique. \\
Now, we demonstrate how this CP uniqueness result can be adapted to our setting. First, consider the matrix $M \in \Rbb^{ps \times q}$ which is obtained by stacking the entries of $T$ as
\begin{align*}
M_{(i-1)s+j,k} = T_{ijk}.
\end{align*}
Then, we have
\begin{align} \label{eq:CP2}
M = (A \odot B)C^\top.
\end{align}
On the other hand, for the 2-persistent topic model with 4 words ($n=2,m=2$), the moment can be written as
\begin{align*}
M^{(2)}_{4}(x) = (A \odot A) \Ebb \bigl[ h h^\top \bigr] (A \odot A)^\top,
\end{align*}
for $A \in \Rbb^{p \times q}$. The following matrix has the same column span of $M^{(2)}_{4}(x)$,
\begin{align*}
M' = (A \odot A) C'^{\top},
\end{align*}
for some full rank matrix $C' \in \Rbb^{q \times q}$.
Our random identifiability result in Theorem \ref{Thm:Identifiability random} provides the uniqueness of $A$ and $C'$, given $M'$, under the size condition $q \leq \bigl( c \frac{p}{2} \bigr)^2$ and the additional degree condition \ref{cond: degree cond}. Note that as discussed in the previous section, this identifiability argument is the same as the unique decomposition of the corresponding tensor.\\
Thus, in equation \eqref{eq:CP2},  by setting  $A=B$ and a full rank square matrix $C$, we obtain the $2$-persistent topic model, under consideration in this paper. Thus, the identifiability results of~\cite{LathauwerSIAM2006} are applicable to our setting, if we assume generic (i.e. fully dense) matrix $A$. However, we incorporate a sparse matrix $A$, and therefore, require different techniques to provide identifiability results. We note that the size bound specified in~\cite{LathauwerSIAM2006} is comparable to the size bound derived in this paper (for random structured matrices), but we have additional degree considerations for identifiability. Analyzing the regime where the uniqueness conditions of~\cite{LathauwerSIAM2006}  are satisfied under sparsity constraints is an interesting question for future investigation.


\end{document}